\pdfoutput=1
\documentclass[11pt]{article}

\usepackage{EMNLP2022}
\usepackage{times}
\usepackage{latexsym}
\usepackage[T1]{fontenc}
\usepackage[utf8]{inputenc}

\usepackage{microtype}
\usepackage{inconsolata}

\usepackage{mathptmx} % <--- better than "times"
\usepackage{latexsym}
\usepackage{xcolor}
\usepackage{stackengine}

\usepackage{adjustbox}
\usepackage{subfigure}

\usepackage{algorithm2e}
\usepackage{setspace}
\usepackage{breqn}
\usepackage{bbm}
\usepackage{dirtytalk}
\usepackage{graphicx}
\usepackage{float}
\usepackage{listings}
\usepackage{booktabs}
\usepackage{caption}
\usepackage{enumitem}
\usepackage{amsmath,stackengine}
\usepackage{cleveref}
\usepackage{babel}
\usepackage{fontawesome}
\stackMath
\title{Detecting Languages Unintelligible to Multilingual Models through Local Structure Probes}
\author{Louis Clouâtre\textsuperscript{1,3}
  Prasanna Parthasarathi\textsuperscript{2}
  Amal Zouaq\textsuperscript{1} and
  Sarath Chandar \textsuperscript{1,3,4} \\
  \textsuperscript{1} Polytechnique Montréal\\
  \textsuperscript{2} Huawei Noah's Ark Lab, Canada\\
  \textsuperscript{3} Quebec Artificial Intelligence Institute (Mila) \\
  \textsuperscript{4} Canada CIFAR AI Chair 
}

\begin{document}
\maketitle

\begin{abstract}
Providing better language tools for low-resource and endangered languages is imperative for equitable growth.
Recent progress with massively multilingual pretrained models has proven surprisingly effective at performing zero-shot transfer to a wide variety of languages.
However, this transfer is not universal, with many languages not currently understood by multilingual approaches.  %\pp{tasks across different families} 
% \pp{With studies estimating that only $72$ of the $7000$ living languages pass the threshold of having a ``small set of labeled datasets'', and intra-family zero-shot transfer among different languages, it becomes necessary to identify which languages are low-resource \emph{and} have a poor language understanding. For, this enables appropriate spending of our limited resources.}
% Of the $7000$ living languages, 
It is estimated that only $72$ languages possess a ``small set of labeled datasets'' on which we could test a model's performance, the vast majority of languages not having the resources available to simply evaluate performances on.
In this work, we attempt to clarify which languages \emph{do} and \emph{do not} currently benefit from such transfer.
% The question of exactly \emph{which} of those languages should we spend some of our limited resources to ameliorate still lacks clarity.
To that end, %\pp{basing on the hypothesis that misplaced characters should affect the sense of a text,} 
we develop a general approach that requires only unlabelled text to detect which languages are not well understood by a cross-lingual model.
% Our approach is derived from the simple hypothesis that if a model's interpretation of ``Apple'' and ``pApel'' are similar, then the model's interpretation of ``Apple'' may be lacking.
Our approach is derived from the hypothesis that if a model's understanding is insensitive to perturbations to text in a language, it is likely to have a limited understanding of that language.
% Our approach is derived from the simple hypothesis that if a model's understanding of a text, once it's had it's characters shuffled, is similar to it's understanding of the original text, t.
We construct a cross-lingual sentence similarity task to evaluate our approach empirically on $350$, primarily low-resource, languages.
%further efforts should be 
% The analysis suggests that efforts should be concentrated in Sub-Saharan African languages, central american languages and languages native to pacific islands to obtain the largest coverage as they are currently the most underserved]
% The analysis suggests [something specific and important for the community].
%Our proposed approach to detect 
% The NLP community has recently emphasized providing better tools for low-resource and endangered languages.
% At the same time, massively multilingual pretrained models have proved surprisingly effective at performing zero-shot transfer to different languages.
% This brings a problem of visibility, exactly which of the 7,000 living languages our current crop of pretrained models perform adequately on, and which languages would it be best to spend some of our limited resources on improving.
% We develop a general approach, requiring only unlabelled text in a target language, that can detect with high precision languages that a cross-lingual model does not understand well.
% We build a cross-lingual sentence similarity dataset covering over 350 languages, providing a large sample of languages to evaluate our method empirically and to provide insights on where to focus our efforts to provide the broadest coverage of languages.
% % Our work demonstrates that languages on which local structure perturbations do not much affect performance

\end{abstract}

\section{Introduction}

% Recent research has shown that neural language models can have an understanding of well-formed English syntax~\citep{gulordava2018colorless,zhang2018language, chrupala-alishahi-2019-correlating, lin-etal-2019-open, belinkov-glass-2019-analysis,liu2019linguistic,jawahar-etal-2019-bert,rogers2020primer}.
% Recent interest in perturbation-based studies suggest that neural models may be insensitive to word-order perturbations~\citep{pham2020out,sinha2021masked,sinha2020unnatural,gupta2021bert,o2021context,Demystifying,ShakingTrees}. 
% Those studies are generally limited to the english language~\citep{pham2020out,sinha2021masked,sinha2020unnatural,gupta2021bert,o2021context,Demystifying} or broadly relies on assumptions regarding the languages~\citep{ShakingTrees}. 
% Assumptions such as the presence of dependency trees, and syntaxic tools.
% Assumptions regarding the specific morphology of the text.
% Even assumptions regarding the presence of whitespaces and words, as understood in the english language, cannot be counted on in a number of low-ressource languages.

% NLP got much better recently, but gains have not been distributed equally to all languages.

Natural Language Processing (NLP) boasts of significant recent successes, largely driven by the introduction of different flavors of pretrained models~\citep{BERT, ALBERT, RoBERTa, GPT-1, GPT-2, GPT-3}.
However, the rewards of those successes have been mostly reaped by high-resource languages.
%The advances in high-resource languages are largely spurred by t
The existence of high-quality benchmarks and metrics, the abundance of readily available high-quality corpora, or the number of researchers speaking the language themselves~\citep{SystematicInequalitiesLT} are significant contributors to the disproportionate advances in high-resource languages.
Although recent improvements in NLP have been shown to extend to several different languages, such as the progress to language understanding by BERT-style models~\citep{ChineseBERT1, FrenchBERT1, FrenchBERT2, ArabicBERT1, BrazilianT5, DutchBERT, SwedeBERT, ItalianBERT, VietBERT}, many of those extensions have been limited to relatively high-resource languages.
Such improvements are often perceived to extend to low-resource languages, but the lack of appropriate benchmarking in those languages curtails our ability to verify such perceptions.
The World Atlas of Language Structures~\citep{wals1, wals2} categorizes over $2600$ languages, and Ethnologue~\citep{EthnologueLanguagesOfTheWorld} estimates that there are currently over $7000$ living languages~\citep{EthnologueReview}; the most popular cross-lingual benchmarks~\citep{XGLUE, XTREME} together cover less than $50$ languages and \citet{StateOfLinguisticDiversityNLP} estimates that only $72$ languages worldwide pass the threshold of having ``a small set of labeled datasets'', which could be used for evaluation.
\textbf{Towards contributing to an equitable society with the development of language technologies, it is imperative that we ensure that no living languages are left behind.}
% \textbf{If the developments in NLP are mostly spurred by research directed at high-resource languages and the improvements found do not translate to all languages, improvements in NLP research may inadvertently disenfranchise speakers of other languages.}
Building automatic and cheap tools to provide better visibility into which languages are not currently well understood by cross-lingual NLP models then becomes essential.
% will help to ensure that proper attention is given to those that most requires it.
%, whether that is gathering data for pretraining or language specific adjustments to modelling.

To determine cheaply if a model understands text in a specific language or not, we first find behaviors that are consistently exhibited by models that do perform well on language understanding tasks.
By finding when those behaviors are not exhibited, we can determine whether a model understands the text or not.
% To determine cheaply if a model understands text in a specific language or not, we would try to find behaviors that are consistently exhibited by models that do perform well on language understanding tasks and look for examples where those behaviors deviate from the norm.
Recent research trends have taken to evaluating well-known natural language understanding (NLU) models on perturbed text~\citep{sinha2020unnatural,sinha2021masked,pham2020out,gupta2021bert,o2021context,ShakingTrees,clouatre-etal-2022-local}.
Such works attempt to distill which aspects of a text are not necessary and which aspects are necessary for language models to understand it by selectively perturbing the text, such as by shuffling the order of words.
% Another perspective on this research trend is that it studies what aspect of a text are \textit{required} for a model to understand it by controlling for all other properties.
It may be possible to use the sensitivity of models to perturbations to properties of text that are found to be essential for NLU as a proxy for model understanding.
As an extreme example, if a model develops the same understanding of a text and the same text with its characters shuffled, it can be hypothesized that its original understanding of the text was limited.
The texts ``I will eat an apple'' and ``ln i plla wat Ieaep'' contain the same characters. 
Yet, we would expect radically different representations of both from a model, assuming it correctly understood the unperturbed version.

% Specifically, \citet{Demystifying} finds that the performance of NLU models in English correlates directly with the degree of local perturbations applied to a text, implying that local structure is essential to building understanding.
% Suppose such a measure holds for many cross-lingual models, languages, and tasks. 
% We may then be able to use the sensitivity to local perturbation of a model on a certain language as a proxy for how much of that language is understood, providing some visibility into specific language failures where previously there was none.
We explore the following research questions and verify their corresponding hypotheses:
\begin{itemize}
    \item \textbf{RQ1: Is there an aspect of text that is universally used by language models that perform well on understanding tasks?}

    \textbf{H1}: We hypothesize that the \textbf{local structure}~\citep{clouatre-etal-2022-local} of text is one such aspect and that the performance of cross-lingual models on language understanding tasks in most languages should be highly sensitive to local structure perturbations. 
    To verify this hypothesis, we use several cross-lingual tasks from popular benchmarks~\citep{XTREME, XGLUE}, on which we perform different local structure-altering perturbations and evaluate several cross-lingual models on said perturbed text.
    % This will provide us with empirical evidence as to whether local structure is consistently relied upon to perform understanding tasks or not.
    
    \item \textbf{RQ2: If there is such a universally relied upon aspect of text, can a model's performance sensitivity to that aspect be used as a proxy for understanding?}
    
    \textbf{H2}: We hypothesize that if such an aspect of text exists, a model that is \textit{in}sensitive to perturbations to that aspect may be inferred to have a limited understanding of the original text.
    To verify this hypothesis, we construct a large-scale cross-lingual sentence representation task covering $350$ languages which we use to measure the language understanding of several models in all the $350$ languages.
    For each model and target language, we measure the sensitivity of perturbations to that aspect.
    We demonstrate that all languages for which our cross-lingual models are less sensitive to perturbations to the local structure of text are also not well understood by those models.
    
    % \item \textbf{RQ3: Language specific further pretraining is a suitable solution for poor performance}
    
    % \textbf{H3}: We hypothesize that further pretraining in a specific language of a cross-lingual model may, with a relatively low requirement for data, address at least some of the performance issues.
\end{itemize}

Our main contributions are:
\begin{itemize}
    \item Across all tested languages, tasks, and models, we find that performance is directly correlated with the amount of local perturbations applied to the text. 
    % \item We confirm that the results of \citet{Demystifying} hold in a multilingual setting and find the conclusions regarding the importance of local structure to be consistent across all tested languages, language families, and text scripts.
    \item We develop the \textbf{monolingual local sensitivity} metric which measures the reliance of a model on the local structure to build text representations, only requiring unlabelled monolingual data.
    \item 
    %From the MTData dataset~\citep{MTData}, we build a cross-lingual sentence retrieval task covering 350 languages-to-English pairs, which lets us cover a large amount of very low-resource languages that are not present in other cross-lingual datasets. 
    On a task covering $350$ languages, we find that languages on which a model has low monolingual local sensitivity always has a poor representation of that language's text.
    % Low monolingual local sensitivity may be a cheap and effective probe to detect language failure modes of cross-lingual models.
    % Rework the analysis if i want to put it back in
    % \item We provide an analysis of languages that are flagged as likely poor performers by our approach.% over several dimensions.
    % We show that those are generally geographically clustered in Central America, Sub-Saharan Africa, and several Pacific Islands.
    % in the three pretrained cross-lingual models tested and find that languages of the Niger-Congo family and Native American languages are likely poor performers that may require additional attention.
\end{itemize}

\section{Related Work}
\label{sec:related-work}

\begin{table*}[ht]
\centering
\begin{adjustbox}{width=0.98\textwidth}
\small
\begin{tabular}{||c c c c c c c c c||} 
 \hline
 \textbf{Languages} & German & Chinese & Spanish & Turkish & Vietnamese & Arabic & Russian & Hindi\\ [0.5ex] 
 \hline\hline
 \textbf{Appearances} & 20 & 20 & 19 & 18 & 17 & 17 & 17 & 16\\ 
 \hline
%  \textbf{Family} & IE: Italic & IE: Germanic & IE: Germanic & IE: Indo-Iranian & IE: Balto-Slavic & Sino-Tibetan & IE: Italic & Afro-Asiatic\\
%  \hline
%  \textbf{Script} & Latin & Latin & Latin & Brahmic & Cyrillic & Chinese characters & Latin & Arabic\\ 
%   \hline
 \textbf{Native Speaker (Millions)} & 95 & 1300 & 493 & 80 & 76 & 400 & 150 & 260\\
 \hline\hline
 \textbf{Languages} & French & Greek & Thai & Bulgarian & Japanese & Korean & Indonesian & Italian\\ [0.5ex] 
 \hline\hline
 \textbf{Appearances} & 16 & 16 & 15 & 13 & 12 & 12 & 12 & 12\\ 
%  \hline
%  \textbf{Family} & IE: Germanic & IE: Italic & IE: Balto-Slavic & Austroasiatic & Japonic & Turkic & IE: Indo-Iranian & IE: Balto-Slavic\\
%  \hline
%  \textbf{Script} & Latin & Latin & Cyrillic & Latin & Kana & Latin & Latin & Latin\\ 
 \hline
 \textbf{Native Speaker (Millions)~\citep{EthnologueLanguagesOfTheWorld}} & 77 & 13 & 28 & 8 & 128 & 80 & 43 & 67\\ [1ex] 
 \hline
\end{tabular}

\end{adjustbox}
\caption{Statistics on languages making the most appearances in the cited cross-lingual performance prediction work.}
\label{tab:language_cross_lingual_val_statistics}
\end{table*}

\paragraph{Cross-Lingual Performance Prediction}
Predicting to what extent cross-lingual models' performances transfer to different languages and tasks has seen a fair amount of interest~\citep{PredictingSuccessTranslation, PredictingPerformanceNLP, limitation_zero_shot_language_transfer, PredictingCrosslingualLinguisticFeatures, PredictingPerformanceMultilingual, PredictingPerformanceNLPFinegrain,  MultiTaskZeroShotPredictingPerformance}.
These works formulate the zero-shot transfer to different languages and tasks as regression problems.
Linguistic features and model-specific features such as the size of the pretraining data and the models' performance in different languages and tasks serve as input.
The performance of the model on a certain type of task and language is then used as the target of the regression.

% has been directed at providing some of the missing visibility by predicting the performance of cross-lingual models on different tasks and languages before training and testing.
% Those work leverage linguistic features of languages, features such as subword overlap between languages and the overall experimental setting to predict a models performance evaluated on an unseen task and language.
% \citet{PredictingSuccessTranslation} predicts the performance of translation between 11 European languages (110 pairs) and shows that the amount of reordering, the morphological
% complexity of the target language and the historical relatedness of the two languages are highly predictive of performance.
% \citet{PredictingPerformanceNLP} builds several features describing both the task and the language, which permits to predict of downstream performance and covers up to 66 languages.
% \citet{PredictingCrosslingualLinguisticFeatures} shows that syntactic features can be highly predictive of cross-lingual transfer for particular tasks.
All those approaches share a few limitations.
They are evaluated on high-resource to medium-resource languages, as those languages all possess supervised learning datasets to be evaluated upon and generally rely upon linguistic features from the World Atlas of Language Structures~\citep{wals1, wals2} which cover less than half of all estimated living languages~\citep{EthnologueReview}.
While those approaches are tremendously valuable for optimizing transfer learning, they provide limited utility in predicting the performance of a model on a very low-resource language.
% A limitation of such studies is the requirement for a supervised learning dataset in a certain language to be able to evaluate the performance of that language.

All cited studies~\citep{PredictingSuccessTranslation, PredictingPerformanceNLP, limitation_zero_shot_language_transfer, PredictingCrosslingualLinguisticFeatures, PredictingPerformanceMultilingual, PredictingPerformanceNLPFinegrain,  MultiTaskZeroShotPredictingPerformance} predicting cross-lingual performance cover a total of $75$ languages.
It may seem like a large selection, but we observe that high-resource languages dominate these works.
By counting the frequencies of appearances of every language used in those works, we find that most of the evaluations were made on some of the world's highest resource languages, in terms of native speakers, as illustrated in Table~\ref{tab:language_cross_lingual_val_statistics}.
Taking an average of the number of native speakers in all languages surveyed, weighted by their appearances in the cited literature, we observe that evaluations were made on languages with, on average, 127 million native speakers.

\paragraph{Text Perturbations and Structure Probing} 
Several text perturbation schemes have been explored in the context of probing model performances.
\citet{sankar-etal-2019-neural} shuffles and reverses utterances and words in a generative dialogue setting, highlighting insensitivity to the order of conversational history.
\citet{pham2020out} shuffles $n$-grams for different values of $n$ , highlighting the insensitivity of pretrained Transformer models.
\citet{sinha2020unnatural} performs perturbations on the position of the words on textual entailment tasks, with the added criterion that all words' positions must have changed.
\citet{ShakingTrees} extend perturbation studies to Swedish and Russian and performs perturbations by shuffling syntactic phrases, rotating sub-trees around the root of the syntactic tree of a sentence, or simply shuffling the words of the text.

These approaches work well to provide insight into many languages with automatic parsing tools or well-developed tokenizers.
However, low-resource languages cannot be assumed to possess those automatic linguistic tools that permit grammatical perturbations.
Language-agnostic tools and measures will need to be prioritized to evaluate the importance of the different aspects of text in low-resource languages.
Priors regarding the form of the text, such as the presence of white-space delimited words, will have to be kept to a minimum.

\citet{clouatre-etal-2022-local} proposes a suite of controllable perturbations on characters, which should be compatible with almost any written language, as well as a metric quantifying perturbations to the \textit{local} structure that measures perturbations on a character-level.
The findings of \citet{clouatre-etal-2022-local} in regards to the ubiquitous nature of \textbf{local sensitivity} as it relates to language understanding and the compatibility of both the metric and perturbations with any text make their work particularly well suited to a massively multilingual setting.

\paragraph{Canine and General Tokenization}
Some of the language scripts used in this work, such as Inuktitut Syllabics, are not covered by the tokenization scheme of most pretrained cross-lingual models such as XLM-R~\citep{XLM} and multilingual-BERT~\citep{BERT}, which rely on a learned vocabulary of subwords~\citep{BPE, wordpiece}.
The Canine model~\citep{Canine} offers a tokenization scheme that covers every Unicode character, allowing it to have representations for scripts that were not part of the pretraining dataset.
This permits us to evaluate low-resource languages in previously unseen scripts that would otherwise have to be ignored.
Has evidence exists that transfer can occur even in languages written in different scripts~\citep{how_m_is_mBERT}, the use of universal tokenization will be necessary to evaluate cross-lingual transfer properly.
Canine also uses character-level tokenization instead of explicitly modeling subwords, which should be more resilient to perturbations to the order of characters and control for the confounder of vocabulary destruction.
% Unlike the ByT5~\citep{ByT5} model, it also boasts computation costs that align with its subword counterpart despite the much longer sequence length used to tokenize into characters directly.
% The pretraining done to mirror the pretraining of mBERT, using the same data, amount of compute and only doing minor modification to the training setup.

\paragraph{Cross-Lingual Sentence Similarity}
Cross-lingual sentence retrieval tasks, such as Tatoeba~\citep{tatoeba}, rely on the presence of language-agnostic sentence embeddings.
By comparing the cosine distance between the embeddings of a text in a target language with the same text in English or another high-resource language, we can obtain a relative idea of the quality of the representations of said target language when compared to its understanding of English.
As models evaluated on English NLU tasks obtain, at times, super-human performances~\citep{GLUE, SuperGLUE}, a model having a similar representation to English sentences in a low-resource language would imply at least \textit{some} level of understanding of that text.
Cross-lingual sentence retrieval is also particularly interesting as, compared to other NLU tasks, obtaining a broad coverage of languages is relatively simple.

% \paragraph{Cross-Lingual Language Understanding (XLU) Datasets}

% \begin{itemize}
%     \item Top datasets
%     \item Even low ressource are kinda high ressource
%     \item thats why extend tatoeba + MTData
% \end{itemize}

\section{Multilingual Local Sensitivity}
To answer \textbf{RQ1}, we borrow some of the perturbation schemes and metrics from \citet{clouatre-etal-2022-local} and apply them to a multilingual setting.
We aim to demonstrate empirically that neural models generally make some use of local structure to perform understanding tasks, irrespective of language.
This can be demonstrated by progressively removing local structure from text through order altering perturbations and observing a similar decline in understanding (as measured by performance metrics) of that text from models.
Such results will motivate using low local structure sensitivity as a proxy for lack of ability to perform language understanding tasks.
We perform those experiments on seven popular cross-lingual tasks covering 44 unique languages.

\subsection{Metric and Perturbations}
The \textbf{CHRF-2} (chrF)~\citep{chrf} metric measures the amount of character bi-gram overlap between a perturbed text and the original text and is used to represent the amount of local structure that has not been perturbed in a text.

\begin{figure}[h!t]
    \centering
    \includegraphics[width=0.95\columnwidth]{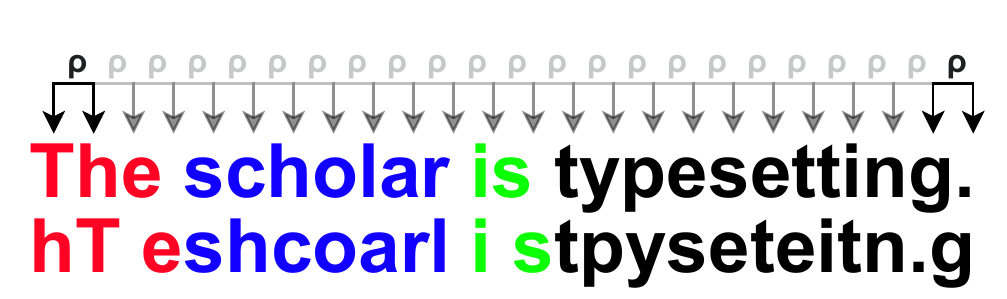}
   
 \caption{From top to bottom: Unperturbed Text, Neighbor Flipping with $\rho=0.5$}
    \label{fig:sample-perturbations}
\end{figure}

We perform perturbations by altering the order of \textbf{characters} present in the text.
This is done by using the \textbf{neighbor flipping}~\citep{clouatre-etal-2022-local} perturbations, which, with a controllable probability $\rho$, flips a character with its neighbor, thus providing an arbitrary amount of local perturbations.
This perturbation is illustrated in \ref{fig:sample-perturbations}.~\footnote{Pseudocode of the perturbation is present in the Appendix~\ref{app:pseudocode_perturbations}}

\subsection{Experimental Details}
All experiments are conducted with the pretrained cross-lingual models Canine-S~\citep{Canine}, XLM-RoBERTa-base (XLM-R)~\citep{XLM} and multilingual-BERT-base-cased (mBERT)~\cite{BERT}.

A total of 7 cross-lingual tasks selected from the most popular cross-lingual benchmarks~\citep{XTREME, XGLUE} covering 44 languages are used for evaluation (see Table~\ref{tab:task_statistics}).~\footnote{Extractive tasks such as extractive QA are not compatible with our perturbations, as the answer would also be perturbed and were not considered.}

\begin{table}[ht]
\centering
\begin{adjustbox}{width=0.95\columnwidth}
\small
\begin{tabular}{||c c c c||} 
 \hline
 \textbf{Task} & $n$ Languages & Task Type & Metric\\ [0.5ex] 
 \hline\hline
 \textbf{PAWS-X} & 7 & Paraphrase Detection & ACC\\ 
 \hline
 \textbf{XNLI} & 15 & NLI & ACC\\ 
 \hline
 \textbf{QAM} & 3 & Text Classification & ACC\\ 
 \hline
 \textbf{QADSM} & 3 & Text Classification & ACC\\ 
 \hline
 \textbf{WPR} & 7 & Page Ranking & nDCG\\ 
 \hline
 \textbf{BUCC} & 5 & Sentence Retrieval & F1\\ 
 \hline
 \textbf{Tatoeba} & 33 & Sentence Retrieval & ACC\\ 
 \hline

\end{tabular}

\end{adjustbox}
\caption{Summary information of the different tasks used.} 
\label{tab:task_statistics}
\end{table}

%~\footnote{Results for XLM-R and mBERT are present in the Appendix~\ref{app:XLM_mBERT}, while results are inline with expectations, the use of vocabulary and the non-coverage of several scripts are heavy confounders.}

The zero-shot cross-lingual setting~\citep{XTREME} is used for all experiments, meaning that the cross-lingual model is finetuned on the English version of the dataset and evaluated without further tuning on all target languages.~\footnote{Detailed training and testing hyperparameters and process are present in the Appendix \ref{app:hyperparameters_training}.}

No finetuning is performed on the cross-lingual sentence retrieval tasks, defaulting to simple cosine similarity of the mean of the final hidden representations of the model for every input token, as described in \citet{XTREME}.

The English version on which the model is finetuned is kept unperturbed, while the target language text on which the model is evaluated goes through several perturbations.
We perform a total of 12 different perturbations on every task and language and obtain their performance, thus evaluating the sensitivity of the target languages to the perturbations.~\footnote{Details perturbations used are present in the Appendix \ref{app:perturbations}}
All models are finetuned on five different random seeds, and all perturbations are performed on five different random seeds, for a total of 25 evaluations for every model on every task, every language present in the tasks, and every perturbation setting.

% Summary information of the tasks can be found in \ref{tab:task_statistics}.~\footnote{As we use all 122 languages in the Tatoeba dataset, which vary from 100 to 1000 possible sentence to retrieve, the F1 score, as used in BUCC, is more appropriate as an evaluation of performance than the accuracy used in the XTREME benchmark.}

\subsection{Results and Discussion}

\begin{figure}[h!t]
    \centering
    \includegraphics[width=0.97\columnwidth]{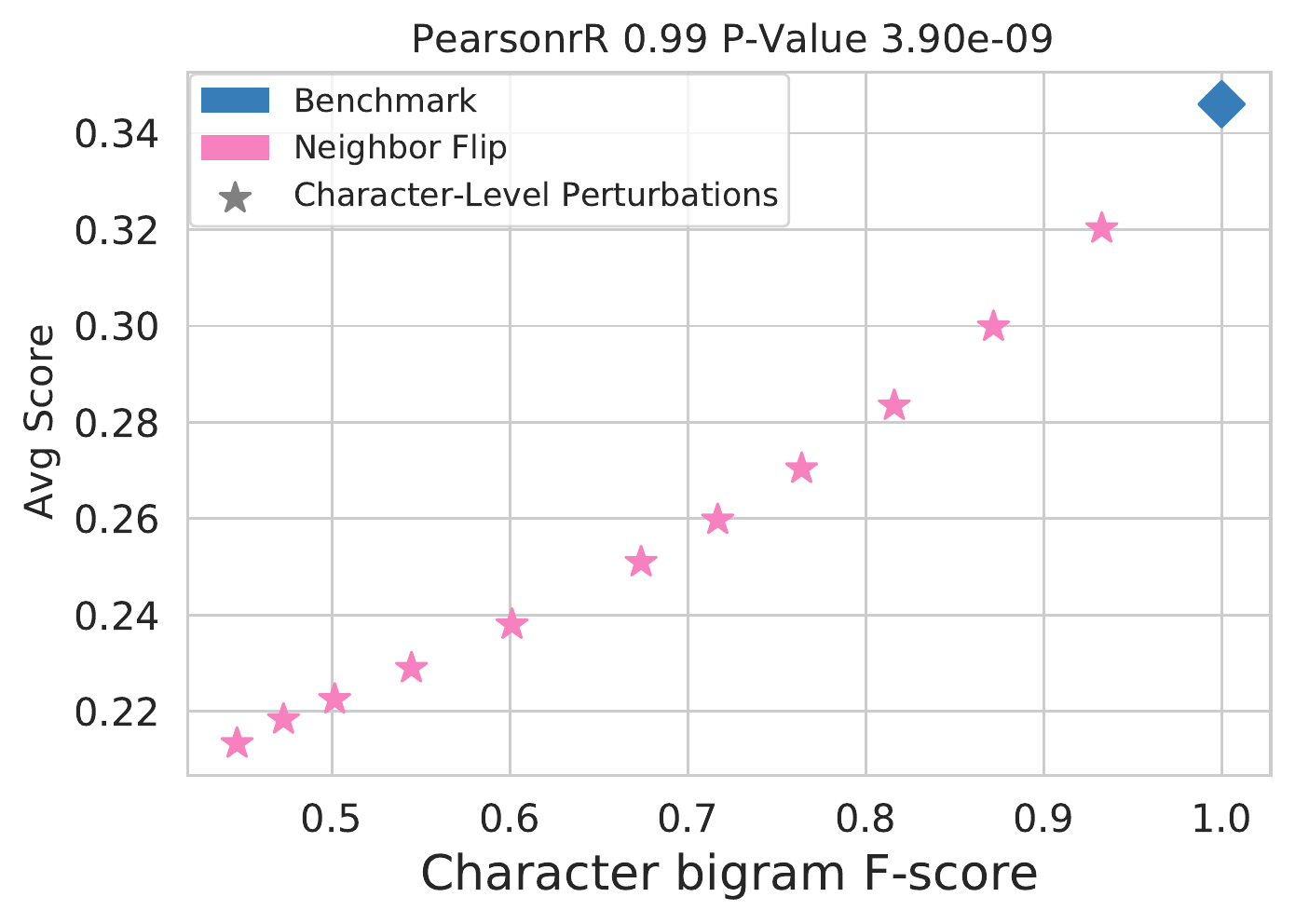}

    \caption{
    Plotted is the relation between local structure perturbations and average performance on all tested datasets and languages, averaged across all models.
    The local sensitivity, measured by the correlation between local perturbations and performance degradations, is reported at the top of the figure.
    }
    \label{fig:global_metrics}
\end{figure}

We observe that, in an aggregate, local structure perturbations almost perfectly correlate with the degradation of the ability of a model to perform language understanding tasks in a cross-lingual setting. 
A Pearson's r of $0.99$ is found between our measure of the perturbations and the performance obtained.
We call this correlation between degradation in performance and the amount of local perturbation the \textbf{local sensitivity}.
Figure~\ref{fig:global_metrics} shows the results averaged across all tasks, all random seeds, and all languages.
We can observe an almost perfect linear relationship between the amount of local structure remaining in the text on which a model is evaluated, as measured by the character bigram F-score, and the average score of our models when evaluated on that text.
% Those results strongly indicate that local structure is always used, at least to some extent, to build understanding from text.

%todo add rank orrelation between performance and local perturbations

% The perturbations to the global structure are shown to be a much poorer explanation for the degradation in performance than the perturbation to the local structure.
% The compression rate is both highly correlated with a model's performance and the local structure, which makes it a potential confounder for the degradation in performance.
% However, the trend in local structure holds with subword-level perturbations, unlike with the compression rate, which is not affected by perturbations to the order of subwords, as well as holding for the vocabulary-free Canine model, as shown in Figure~\ref{fig:models-metric-correlation}.

% This makes it more likely that the cause for the degradation in performance is the local structure perturbation, the destruction of the vocabulary being incidental.

%todo add rank orrelation between performance and local perturbations

\begin{figure}[ht]
    \centering
    \includegraphics[width=0.95\columnwidth]{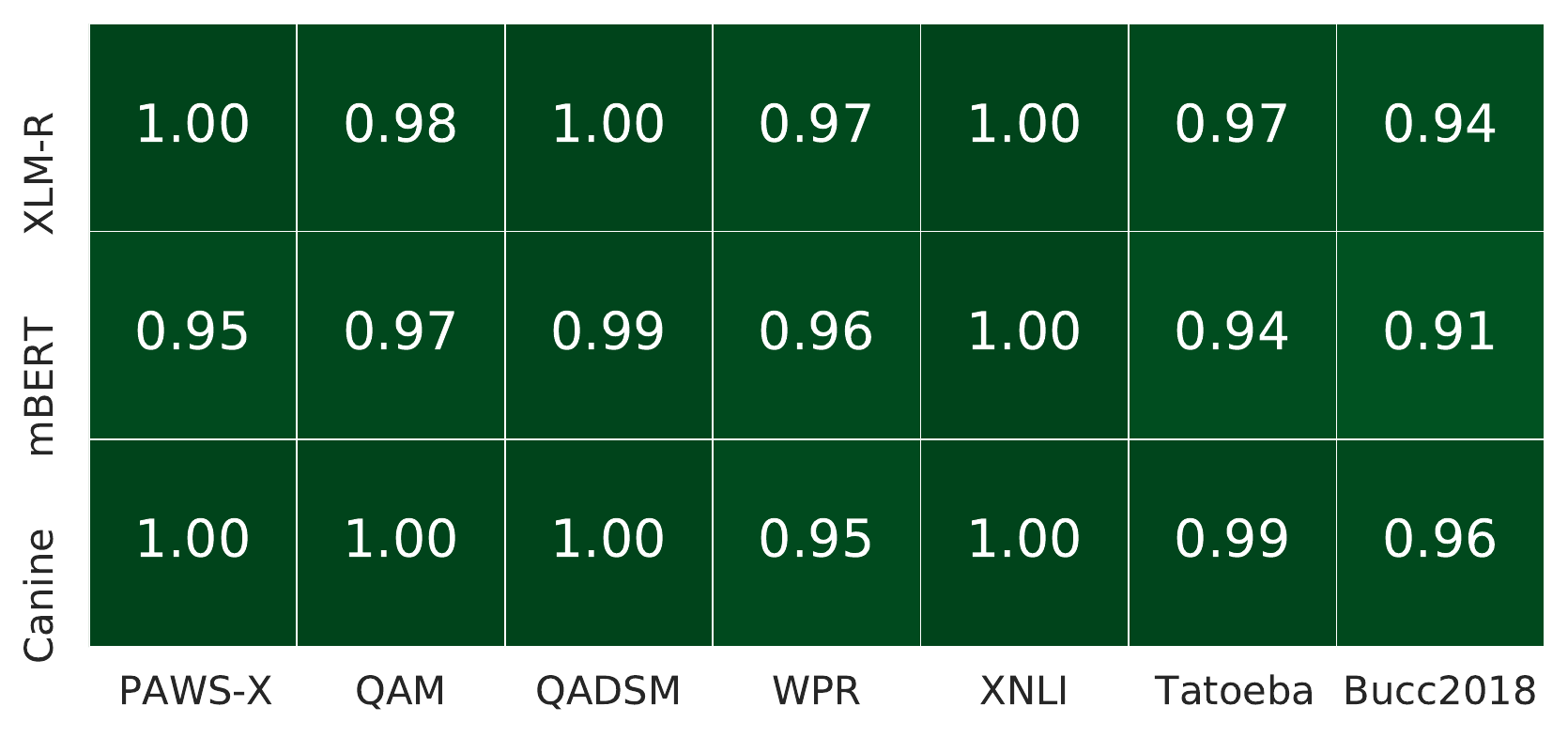}
    \caption{Local sensitivity matrix of the different models tested on the various tasks averaged across all random seeds.
    The higher the value, the more sensitive a model's performance is to perturbations to local structure.}
    \label{fig:task-metric-correlation}
\end{figure}

The local sensitivities of the different models on the various tasks are also very consistent, performances being either perfectly or highly correlated to the amount of local structure remaining, as pictured in Figure~\ref{fig:task-metric-correlation}.
This is consistent across all models, including the tokenization-free Canine, which lets us control for the vocabulary destruction brought by perturbing the order of characters.
% This trend of extremely high correlation between performance and perturbations also holds when grouping results by script and language family, as shown in Figure~\ref{fig:family-metric-correlation} and Figure~\ref{fig:script-metric-correlation}.

% \begin{figure}[ht]
%     \centering
%     \includegraphics[width=0.99\columnwidth]{img/family_correlation_score.pdf}
%     \caption{Local sensitivity matrix between the different languages families with at least 3 tested languages in our tasks, averaged across all tasks and models.}
%     \label{fig:family-metric-correlation}
% \end{figure}

% \begin{figure}[ht]
%     \centering
%     \includegraphics[width=0.99\columnwidth]{img/script_correlation_score.pdf}
%     \caption{Local sensitivity matrix between the different scripts with at least 3 tested languages in our tasks, averaged across all tasks and models.}
%     \label{fig:script-metric-correlation}
% \end{figure}

\begin{figure}[ht]
    \centering
    \includegraphics[width=0.99\columnwidth]{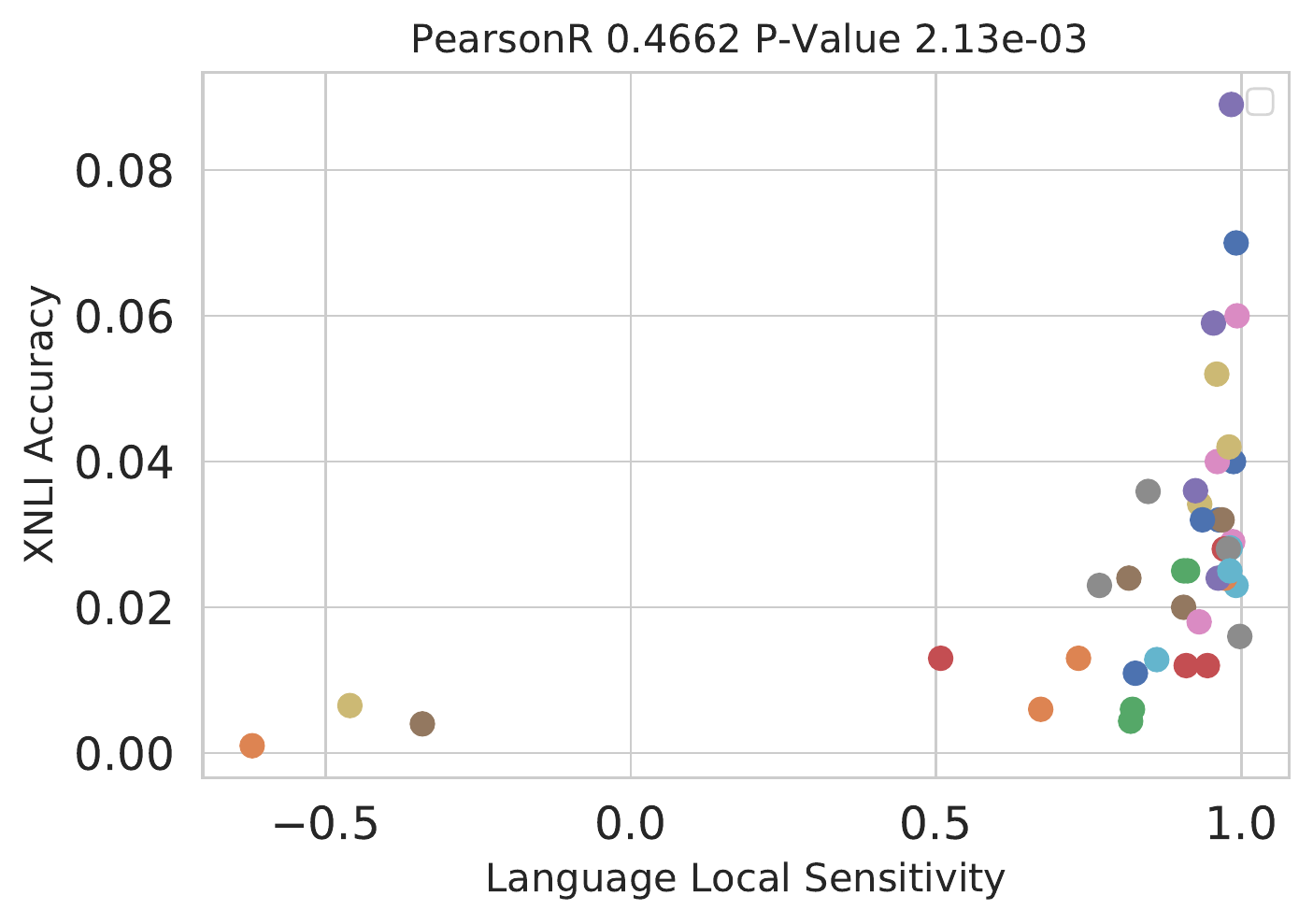}
    \caption{Plotted are the individual language's local sensitivity plotted against their performance on the unperturbed text on the XNLI task, averaged across all models.}
    \label{fig:corr-perf-xnli}
\end{figure}

Finally, we can observe whether or not languages with lower local sensitivity tend to underperform their locally sensitive counterparts. 
In Figure~\ref{fig:corr-perf-xnli}, we observe that while high local sensitivity does not guarantee good performance, none of the languages that posses low local sensitivity do much better then chance on the task of Natural Language Inference. 
Those results are consistent across all tasks and present in Appendix~\ref{app:results_cross_lingual}.

% todo, bad language curve

From our results, we cannot find a dimension in which a model's performance is not extremely sensitive to local structure perturbations, lending credence that local structure is an aspect of text that is always, at least, relied upon to perform understanding tasks.
Those results support \textbf{H1}, demonstrating that it is likely that language models universally make some use of local structure to perform understanding tasks, irrespective of language, task, or the specific pretrained model.
Further, in our tested tasks, languages on which a model has low local sensitivity tend to underperform those with high local sensitivity.

\section{Low Monolingual Local Sensitivity as a Proxy for Lack of Understanding}

% With the extension to \citet{clouatre-etal-2022-local} to a cross-lingual setting, we ca
This section explores using an insensitivity to local perturbations as a proxy for lack of understanding to address \textbf{RQ2}.
To find a proxy for understanding that will provide greater visibility in the performance of very low-resource languages where evaluation is not possible, we cannot measure the local sensitivity by evaluating a language on a labeled task.
Therefore, we will explore \textbf{monolingual local sensitivity} as a proxy for lack of language understanding, with unlabelled monolingual data in the target language as its only requirement.

\subsection{Monolingual Local Sensitivity}\label{sec:monolingual_local}
We previously defined local sensitivity as the correlation between the degradation of performance of a model on a task and the local perturbations applied to its text.
To calculate the local sensitivity of a model on a specific task, we evaluate the model's performance on that task with all 12 of our perturbations and calculate the Pearson's r between the performance on the perturbed text and the local structure as measured by CHRF-2.
However, this process has the limitation of requiring a labeled dataset on which to evaluate performance.

To obtain a measure of local sensitivity while bypassing the requirement for a supervised learning dataset, we turn to the \textbf{monolingual local sensitivity}.
First, we build a corpus in the target language containing 1000 unique texts.
We then formulate the problem as a sentence similarity between two copies of the same corpus, initially resulting in a perfect similarity between sentence pairs.
We apply our perturbations to one copy of the corpus while keeping the other copy unperturbed.
As more of the local structure is destroyed, the representation of the different pieces of text should also drift apart, assuming that the model considers local structure.
We can then obtain a measure of local sensitivity based on the task of sentence retrieval between the same corpus, one of which is perturbed.
A toy example comparing cross-lingual sentence similarity and monolingual sentence similarity is pictured in Figure~\ref{fig:monolingual_perturbations}.
% ~\footnote{Examples illustrating how the monolingual local sensitivity is calculated are present in the Appendix \ref{app:monolingual}.}

\begin{figure*}[ht]
    \centering
    \includegraphics[width=0.95\textwidth]{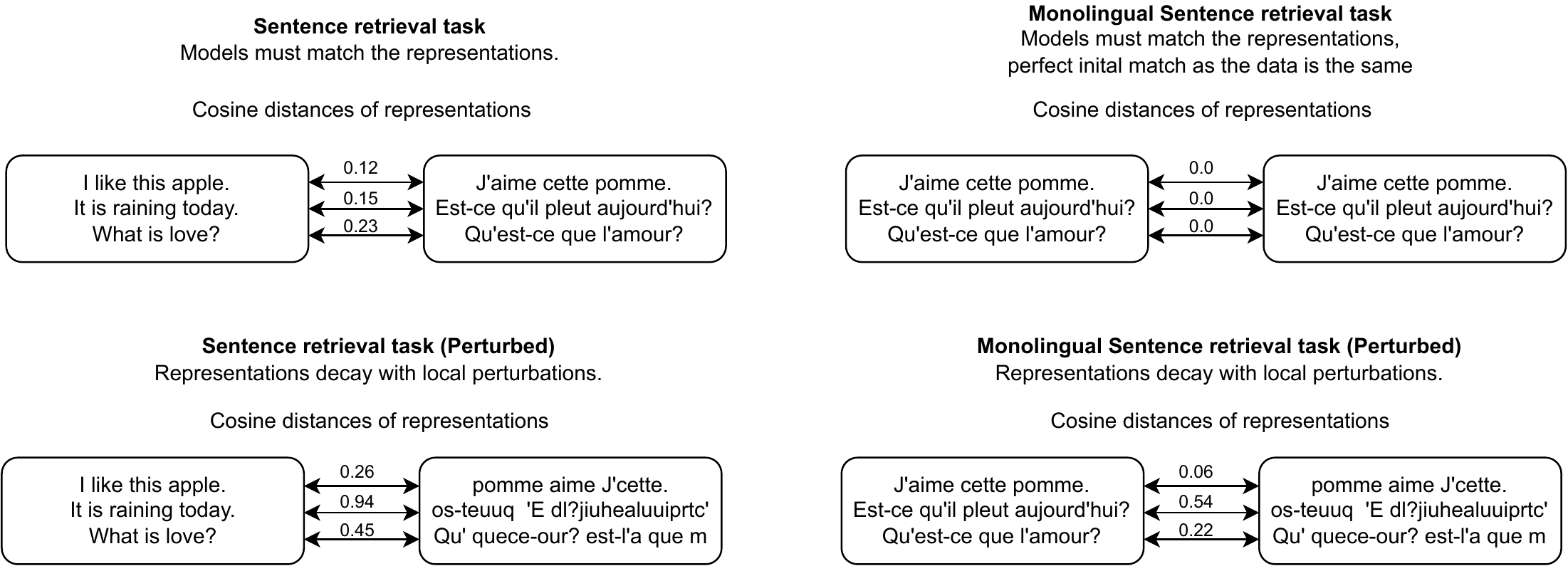}
    \caption{Toy example of sentence similarity and monolingual sentence similarity with and without perturbations.}
    \label{fig:monolingual_perturbations}
\end{figure*}

% \begin{figure}[ht]
%     \centering
%     \fbox{\begin{minipage}{0.9\columnwidth}
%     \setstretch{1.5}
%     \parbox{\columnwidth}{
%     	    \centering
    
%             \large
%             Placeholder, should illustrate monolingual local sensitivity
%             % \textcolor{red}{The} \textcolor{blue}{scholar} \textcolor{green}{is} \textcolor{black}{typesetting}.
            
%             % \textcolor{blue}{scholar}
%             % \textcolor{black}{typesetting}
%             % \textcolor{green}{is}
%             % \textcolor{red}{The}.
%         }
%      \end{minipage}
%      }
%     \caption{Example of a text used to calculate monolingual local sensitivity.}
%     \label{fig:sample-perturbations_word_full_shuffle}
% \end{figure}

% \begin{itemize}
%     \item Monolingual local sensitivity involves a sentence retrieval between two identical corpus
%     \item With no perturbations, the score of a model is always perfect, as it looks for the nearest identical neighbour
%     \item 
% \end{itemize}

\subsection{MTData Sentence Retrieval}
We first require a simple task covering many low-resource languages to evaluate low monolingual local sensitivity as an indicator of lack of understanding in a meaningful way.
From the MTData~\citep{MTData} dataset, which is composed of millions of sentence pairs between English and over 500 target languages, we build an English-to-language cross-lingual sentence similarity task covering 350 different language-to-English pairs containing 1000 text pairs per language.
The dataset is built using the same process and filtering as was used to construct the Tatoeba cross-lingual sentence similarity dataset~\citep{tatoeba}.~\footnote{Specific details, dataset statistics, and evaluation methods are expanded upon in the Appendix~\ref{app:mtdata}.}
We will use the normalized cosine similarity between the sentence representation and its target representation to evaluate performance.
We normalize by removing the mean and scaling it by the standard deviation of cosine similarity between the text and all other potential texts.
This evaluation metric should control for the different models' behaviors, the different quality of corpora for the different languages, and the diversity of examples for every language, making comparisons more uniform than a simple cosine distance and less sparse than an absolute hit rate.
% In essence, we score our model on how many standard deviations higher is the cosine similarity between the target and the prediction when compared to its cosine similarity with every other potential target text.
Under this scoring system, a score of $1.0$ would mean that the representation of a text with its counterpart would be $1.0$ standard deviation closer than its distance to all other texts, as measured by the cosine distance.
We will refer to this metric as the \textbf{similarity Z-Score}.

% \begin{itemize}
%     \item Reframe since problem with current setting
%     \item Basically, we build the dataset to be able to test, but only same language are considered
%     \item This is cleaner as an explanation, doesnt muddy the waters with the other example, and should be much more stable
% \end{itemize}
\subsection{Results and Discussion}
From our MTData cross-lingual sentence similarity task, we can obtain and compare two measures for the 350 languages.

The first is the model's performance on the task of cross-lingual similarity between an English representation, which is assumed to be of reasonable quality as the model performs well on English understanding tasks, and a target language representation, as measured by the similarity Z-Score.
The closer the representation target language's text to its English representation, the closer the abilities of the model to represent that language are to the ability of the model to represent English.

The second is the monolingual local sensitivity, which we obtain by performing sentence retrieval using two copies of the target language side of the MTData cross-lingual sentence retrieval dataset, as illustrated in the left side of Figure~\ref{fig:monolingual_perturbations}.

\begin{figure}[h!t]
    \centering
    \includegraphics[width=0.97\columnwidth]{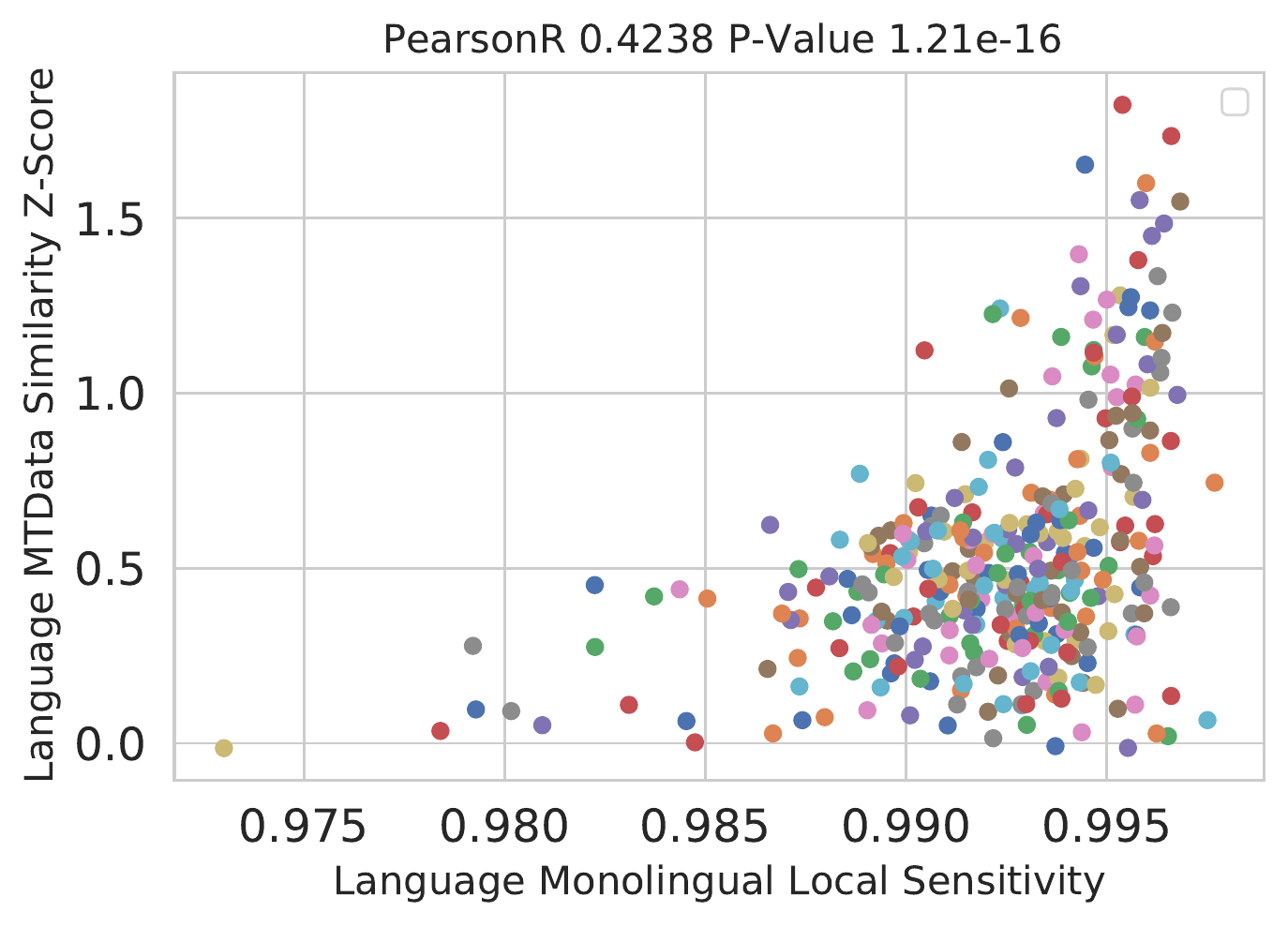}

    \caption{Scatter plot of all $350$ languages comparing their degree of monolingual local sensitivity and their similarity Z-Score on the MTData cross-lingual sentence similarity task.
    Left to right is less sensitive to more sensitive to monolingual local perturbations.
    The Pearson's r between a language's monolingual local sensitivity and its cross-lingual similarity Z-Score is reported.}
    \label{fig:mtdata_lang_lang_unperturbed}
\end{figure}

\begin{figure*}[ht]
    \centering
    \includegraphics[width=0.97\textwidth]{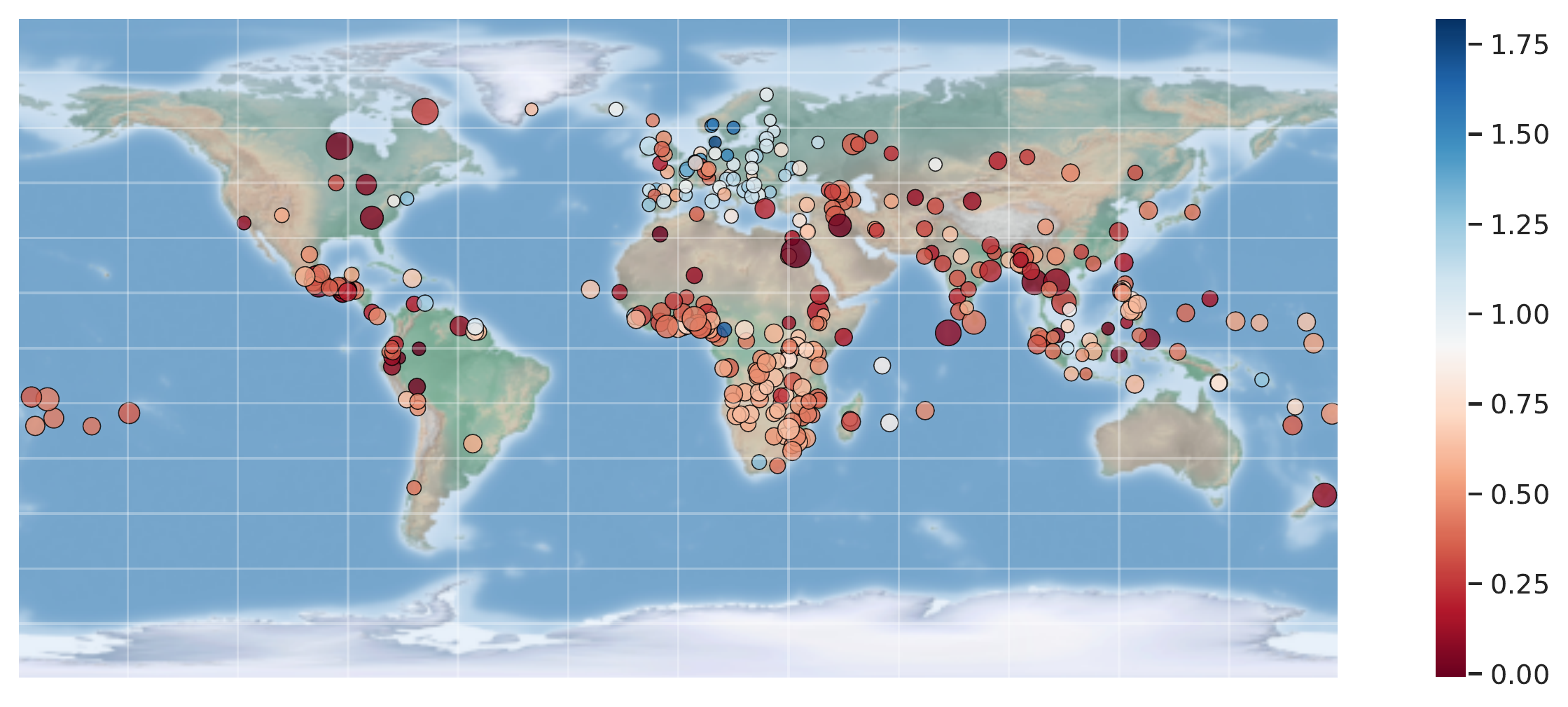}\hfill

    \caption{
    All $350$ languages are plotted on their estimated geographical centers.
    The color of the dots are scaled by the cross-lingual similarity Z-Score while their size are scaled by how low the monolingual local sensitivity is for that language.   
    Large red dots represent languages that were both has poor performance and low monolingual local sensitivity. 
    Small blue dots represent languages that both had good performance and high monolingual local sensitivity.}
    \label{fig:world_map_correlations}
\end{figure*}

We compare the performance of our pretrained models against the monolingual local sensitivity of all $350$ tested languages, pictured in Figure~\ref{fig:mtdata_lang_lang_unperturbed}.
Languages with high local sensitivity may often have poor unperturbed performance, meaning that relying on local structure does not imply good language understanding.
The opposite, however, seems broadly true.
Specifically, languages with low monolingual local sensitivity have universally poor unperturbed performance.
To build representations roughly in line with the quality of an English representation, a model must rely, at least somewhat, on that text's local structure.

All languages that obtained a monolingual local sensitivity of under $0.99$ did not have representations that were very close to their English counterparts.
Assuming normality, a similarity Z-Score of $0.8$ implies that over $21\%$ of representations outputted by the model for that language were closer to the representation of the  English counterpart than the target text pair.
None of the languages with monolingual local sensitivity under $0.99$ clear that hurdle.
Surprisingly, from the $350$ languages surveyed, only a few could truly be said not to be understood by the models.
The probability of having an average score of even $0.10$ on this task through a random process is vanishingly small, and only $23$ of the $350$ languages do not cross that threshold, an encouraging result for the current multilingual pretraining approaches.

To provide greater context on those results, we have plotted all $350$ surveyed languages on their estimated geographical centers~\citep{wals1, wals2}, in Figure~\ref{fig:world_map_correlations}.
We can observe several languages that both underperform, as indicated by the color, and are predicted to underperform, as indicated by the size of the circle.
Further analysis of our results is provided in the Appendix~\ref{app:low_ressource}.

% \begin{figure*}[h!t]
%     \centering
%     \includegraphics[width=0.33\textwidth]{img/monolingual_perf_Language.pdf}\hfill
%     \includegraphics[width=0.33\textwidth]{img/monolingual_perf_Family.pdf}\hfill
%     \includegraphics[width=0.33\textwidth]{img/monolingual_perf_Script.pdf}

%     \caption{Scatter plots of all 350 languages comparing their degree of monolingual local sensitivity and their unperturbed performance on the MTData cross-lingual sentence similarity task.
%     From left to right, individual languages, languages grouped by language families and languages grouped by scripts.
%     The higher the value the more "understanding" of that language a model has.
%     The pearson correlation is reported.}
%     \label{fig:mtdata_lang_lang_unperturbed}
% \end{figure*}

From our results, we find that a low monolingual sensitivity is indicative of a limited ability to represent text. 
Those results support \textbf{H2}, demonstrating that it is likely that a model's inability to properly represent a certain language can be detected through monolingual local structure probes.

\section{Limitations and Ethical Considerations}

% We have developed an approach that detects, with high precision, languages that are likely not well understood by cross-lingual models.
There are several limitations to our approach that may have an ethical impact.

The first one is the poor recall.
While languages with low monolingual local sensitivity, from our experiments, always are poor performers, many poor performers are also sensitive to local perturbations.
Our approach can successfully find some languages that do require additional attention but will miss many other languages.
If we rely on automatic tools to detect where to put efforts, there is a possibility that no efforts are put on languages that are not detected by those tools.

The second one is the data requirement.
Obtaining a sufficient sample of text to calculate monolingual local sensitivity for some low-resource languages may still be too high of a hurdle.
Some living languages, especially those from oral traditions, may have a limited pool of written text available.

It is crucial that if we use automatic tools to detect which languages requires further efforts, we do not forget of the languages that might not be detected or are incompatible with those tools.

\section{Conclusion}

Regardless of the language, task, or model used, the use of local structure seems to be relied upon by neural models to build an understanding of text.
Local structure sensitivity does not seem to be an artifact of the English language and broadly applies to written text in most languages.

% A major exception being when grammatical cues are essential to complete the task, such as in CoLA or PAWS-X.
% We explore this phenomenon further, showing that it is not that local structure is not essential, but that perturbations to word order are sufficient to destroy performance completely.

% Languages using Chinese characters as their script also deviate from the norm.
% We discuss the peculiarities of Chinese script tokenization, their smallest units being semantically much richer than latin characters, which we identify as the cause for the deviation in the overall trend.
% A potential follow-up would be to explore datasets built with the Pinyin language, the romanized version of standard Chinese, to isolate the impact of the script on the results.

We explore monolingual local sensitivity to automatically detect unintelligible languages to cross-lingual models, the only requirement being access to monolingual unlabelled text.
If local structure is essential to building understanding, not relying on the local structure would imply a limited understanding.
We demonstrate a high correlation between monolingual local sensitivity and the ability of a model to perform cross-lingual sentence similarity in $350$ diverse languages.
Specifically, all languages with low monolingual local sensitivity performed poorly on that task.
Those results indicate that with the measure of monolingual local sensitivity alone, it is possible to estimate the performance of a certain language on a certain model without access to any supervised learning datasets.

Our contribution will be useful to direct further efforts, such as unlabelled data gathering for pretraining, to expand the coverage of cross-lingual models in the most efficient way.

\section*{Acknowledgements}
This research has been funded by the NSERC Discovery Grant Program.

\newpage
\bibliography{anthology,custom}

\begin{thebibliography}{48}
\expandafter\ifx\csname natexlab\endcsname\relax\def\natexlab#1{#1}\fi

\bibitem[{Ahuja et~al.(2022)Ahuja, Kumar, Dandapat, and
  Choudhury}]{MultiTaskZeroShotPredictingPerformance}
Kabir Ahuja, Shanu Kumar, Sandipan Dandapat, and Monojit Choudhury. 2022.
\newblock \href {https://aclanthology.org/2022.acl-long.374} {Multi task
  learning for zero shot performance prediction of multilingual models}.
\newblock In \emph{Proceedings of the 60th Annual Meeting of the Association
  for Computational Linguistics (Volume 1: Long Papers)}, pages 5454--5467,
  Dublin, Ireland. Association for Computational Linguistics.

\bibitem[{Antoun et~al.(2020)Antoun, Baly, and Hajj}]{ArabicBERT1}
Wissam Antoun, Fady Baly, and Hazem~M. Hajj. 2020.
\newblock Arabert: Transformer-based model for arabic language understanding.
\newblock \emph{ArXiv}, abs/2003.00104.

\bibitem[{Artetxe and Schwenk(2018)}]{tatoeba}
Mikel Artetxe and Holger Schwenk. 2018.
\newblock \href {http://arxiv.org/abs/1812.10464} {Massively multilingual
  sentence embeddings for zero-shot cross-lingual transfer and beyond}.
\newblock \emph{CoRR}, abs/1812.10464.

\bibitem[{Birch et~al.(2008)Birch, Osborne, and
  Koehn}]{PredictingSuccessTranslation}
Alexandra Birch, Miles Osborne, and Philipp Koehn. 2008.
\newblock \href {https://aclanthology.org/D08-1078} {Predicting success in
  machine translation}.
\newblock In \emph{Proceedings of the 2008 Conference on Empirical Methods in
  Natural Language Processing}, pages 745--754, Honolulu, Hawaii. Association
  for Computational Linguistics.

\bibitem[{Blasi et~al.(2022)Blasi, Anastasopoulos, and
  Neubig}]{SystematicInequalitiesLT}
Damian Blasi, Antonios Anastasopoulos, and Graham Neubig. 2022.
\newblock \href {https://aclanthology.org/2022.acl-long.376} {Systematic
  inequalities in language technology performance across the world{'}s
  languages}.
\newblock In \emph{Proceedings of the 60th Annual Meeting of the Association
  for Computational Linguistics (Volume 1: Long Papers)}, pages 5486--5505,
  Dublin, Ireland. Association for Computational Linguistics.

\bibitem[{Brown et~al.(2020)Brown, Mann, Ryder, Subbiah, Kaplan, Dhariwal,
  Neelakantan, Shyam, Sastry, Askell, Agarwal, Herbert-Voss, Krueger, Henighan,
  Child, Ramesh, Ziegler, Wu, Winter, Hesse, Chen, Sigler, Litwin, Gray, Chess,
  Clark, Berner, McCandlish, Radford, Sutskever, and Amodei}]{GPT-3}
Tom~B. Brown, Benjamin Mann, Nick Ryder, Melanie Subbiah, Jared Kaplan,
  Prafulla Dhariwal, Arvind Neelakantan, Pranav Shyam, Girish Sastry, Amanda
  Askell, Sandhini Agarwal, Ariel Herbert-Voss, Gretchen Krueger, Tom Henighan,
  Rewon Child, Aditya Ramesh, Daniel~M. Ziegler, Jeffrey Wu, Clemens Winter,
  Christopher Hesse, Mark Chen, Eric Sigler, Mateusz Litwin, Scott Gray,
  Benjamin Chess, Jack Clark, Christopher Berner, Sam McCandlish, Alec Radford,
  Ilya Sutskever, and Dario Amodei. 2020.
\newblock \href {https://doi.org/10.48550/ARXIV.2005.14165} {Language models
  are few-shot learners}.

\bibitem[{Carmo et~al.(2020)Carmo, Piau, Campiotti, Nogueira, and
  de~Alencar~Lotufo}]{BrazilianT5}
Diedre Carmo, Marcos Piau, Israel Campiotti, Rodrigo Nogueira, and Roberto
  de~Alencar~Lotufo. 2020.
\newblock \href {http://arxiv.org/abs/2008.09144} {{PTT5:} pretraining and
  validating the {T5} model on brazilian portuguese data}.
\newblock \emph{CoRR}, abs/2008.09144.

\bibitem[{Clark et~al.(2021)Clark, Garrette, Turc, and Wieting}]{Canine}
Jonathan~H. Clark, Dan Garrette, Iulia Turc, and John Wieting. 2021.
\newblock \href {https://doi.org/10.48550/ARXIV.2103.06874} {Canine:
  Pre-training an efficient tokenization-free encoder for language
  representation}.

\bibitem[{Clouatre et~al.(2022)Clouatre, Parthasarathi, Zouaq, and
  Chandar}]{clouatre-etal-2022-local}
Louis Clouatre, Prasanna Parthasarathi, Amal Zouaq, and Sarath Chandar. 2022.
\newblock \href {https://doi.org/10.18653/v1/2022.findings-acl.293} {Local
  structure matters most: Perturbation study in {NLU}}.
\newblock In \emph{Findings of the Association for Computational Linguistics:
  ACL 2022}, pages 3712--3731, Dublin, Ireland. Association for Computational
  Linguistics.

\bibitem[{Cui et~al.(2019)Cui, Che, Liu, Qin, Yang, Wang, and
  Hu}]{ChineseBERT1}
Yiming Cui, Wanxiang Che, Ting Liu, Bing Qin, Ziqing Yang, Shijin Wang, and
  Guoping Hu. 2019.
\newblock \href {http://arxiv.org/abs/1906.08101} {Pre-training with whole word
  masking for chinese {BERT}}.
\newblock \emph{CoRR}, abs/1906.08101.

\bibitem[{de~Vries et~al.(2019)de~Vries, van Cranenburgh, Bisazza, Caselli, van
  Noord, and Nissim}]{DutchBERT}
Wietse de~Vries, Andreas van Cranenburgh, Arianna Bisazza, Tommaso Caselli,
  Gertjan van Noord, and Malvina Nissim. 2019.
\newblock \href {http://arxiv.org/abs/1912.09582} {Bertje: {A} dutch {BERT}
  model}.
\newblock \emph{CoRR}, abs/1912.09582.

\bibitem[{Devlin et~al.(2019)Devlin, Chang, Lee, and Toutanova}]{BERT}
J.~Devlin, Ming-Wei Chang, Kenton Lee, and Kristina Toutanova. 2019.
\newblock Bert: Pre-training of deep bidirectional transformers for language
  understanding.
\newblock In \emph{NAACL}.

\bibitem[{Dolicki and
  Spanakis(2021)}]{PredictingCrosslingualLinguisticFeatures}
Blazej Dolicki and Gerasimos Spanakis. 2021.
\newblock \href {http://arxiv.org/abs/2105.05975} {Analysing the impact of
  linguistic features on cross-lingual transfer}.
\newblock \emph{CoRR}, abs/2105.05975.

\bibitem[{Dryer and Haspelmath(2013)}]{wals2}
Matthew~S. Dryer and Martin Haspelmath, editors. 2013.
\newblock \href {https://wals.info/} {\emph{WALS Online}}.
\newblock Max Planck Institute for Evolutionary Anthropology, Leipzig.

\bibitem[{Eberhard et~al.(2022)Eberhard, Simons, and
  Fennig}]{EthnologueLanguagesOfTheWorld}
David~M. Eberhard, Gary~F. Simons, and Charles~D. Fennig. 2022.
\newblock \emph{Ethnologue: Languages of the World}, twenty-fifth edition.
\newblock SIL International.

\bibitem[{Gowda et~al.(2021)Gowda, Zhang, Mattmann, and May}]{MTData}
Thamme Gowda, Zhao Zhang, Chris~A Mattmann, and Jonathan May. 2021.
\newblock \href {https://doi.org/10.48550/ARXIV.2104.00290} {Many-to-english
  machine translation tools, data, and pretrained models}.

\bibitem[{Gupta et~al.(2021)Gupta, Kvernadze, and Srikumar}]{gupta2021bert}
Ashim Gupta, Giorgi Kvernadze, and Vivek Srikumar. 2021.
\newblock Bert \& family eat word salad: Experiments with text understanding.
\newblock \emph{arXiv preprint arXiv:2101.03453}.

\bibitem[{Hammarstr{\"o}m(2015)}]{EthnologueReview}
Harald Hammarstr{\"o}m. 2015.
\newblock Ethnologue 16/17/18th editions: A comprehensive review.
\newblock \emph{Language}, 91:723 -- 737.

\bibitem[{Haspelmath et~al.(2014)Haspelmath, Bibiko, and Schmidt}]{wals1}
Martin Haspelmath, Hans-Jörg Bibiko, and Claudia Schmidt. 2014.
\newblock \emph{The World Atlas of Language Structures}.
\newblock Oxford University Press.

\bibitem[{Hu et~al.(2020)Hu, Ruder, Siddhant, Neubig, Firat, and
  Johnson}]{XTREME}
Junjie Hu, Sebastian Ruder, Aditya Siddhant, Graham Neubig, Orhan Firat, and
  Melvin Johnson. 2020.
\newblock \href {http://arxiv.org/abs/2003.11080} {{XTREME:} {A} massively
  multilingual multi-task benchmark for evaluating cross-lingual
  generalization}.

\bibitem[{Joshi et~al.(2020)Joshi, Santy, Budhiraja, Bali, and
  Choudhury}]{StateOfLinguisticDiversityNLP}
Pratik Joshi, Sebastin Santy, Amar Budhiraja, Kalika Bali, and Monojit
  Choudhury. 2020.
\newblock \href {https://doi.org/10.18653/v1/2020.acl-main.560} {The state and
  fate of linguistic diversity and inclusion in the {NLP} world}.
\newblock In \emph{Proceedings of the 58th Annual Meeting of the Association
  for Computational Linguistics}, pages 6282--6293, Online. Association for
  Computational Linguistics.

\bibitem[{Lample and Conneau(2019)}]{XLM}
Guillaume Lample and Alexis Conneau. 2019.
\newblock \href {http://arxiv.org/abs/1901.07291} {Cross-lingual language model
  pretraining}.
\newblock \emph{CoRR}, abs/1901.07291.

\bibitem[{Lan et~al.(2019)Lan, Chen, Goodman, Gimpel, Sharma, and
  Soricut}]{ALBERT}
Zhenzhong Lan, Mingda Chen, Sebastian Goodman, Kevin Gimpel, Piyush Sharma, and
  Radu Soricut. 2019.
\newblock \href {http://arxiv.org/abs/1909.11942} {{ALBERT:} {A} lite {BERT}
  for self-supervised learning of language representations}.
\newblock \emph{CoRR}, abs/1909.11942.

\bibitem[{Lauscher et~al.(2020)Lauscher, Ravishankar, Vuli{\'c}, and
  Glava{\v{s}}}]{limitation_zero_shot_language_transfer}
Anne Lauscher, Vinit Ravishankar, Ivan Vuli{\'c}, and Goran Glava{\v{s}}. 2020.
\newblock \href {https://doi.org/10.18653/v1/2020.emnlp-main.363} {From zero to
  hero: {O}n the limitations of zero-shot language transfer with multilingual
  {T}ransformers}.
\newblock In \emph{Proceedings of the 2020 Conference on Empirical Methods in
  Natural Language Processing (EMNLP)}, pages 4483--4499, Online. Association
  for Computational Linguistics.

\bibitem[{Le et~al.(2019)Le, Vial, Frej, Segonne, Coavoux, Lecouteux, Allauzen,
  Crabb{\'{e}}, Besacier, and Schwab}]{FrenchBERT1}
Hang Le, Lo{\"{\i}}c Vial, Jibril Frej, Vincent Segonne, Maximin Coavoux,
  Benjamin Lecouteux, Alexandre Allauzen, Beno{\^{\i}}t Crabb{\'{e}}, Laurent
  Besacier, and Didier Schwab. 2019.
\newblock \href {http://arxiv.org/abs/1912.05372} {Flaubert: Unsupervised
  language model pre-training for french}.
\newblock \emph{CoRR}, abs/1912.05372.

\bibitem[{Liang et~al.(2020)Liang, Duan, Gong, Wu, Guo, Qi, Gong, Shou, Jiang,
  Cao, Fan, Zhang, Agrawal, Cui, Wei, Bharti, Qiao, Chen, Wu, Liu, Yang,
  Majumder, and Zhou}]{XGLUE}
Yaobo Liang, Nan Duan, Yeyun Gong, Ning Wu, Fenfei Guo, Weizhen Qi, Ming Gong,
  Linjun Shou, Daxin Jiang, Guihong Cao, Xiaodong Fan, Bruce Zhang, Rahul
  Agrawal, Edward Cui, Sining Wei, Taroon Bharti, Ying Qiao, Jiun{-}Hung Chen,
  Winnie Wu, Shuguang Liu, Fan Yang, Rangan Majumder, and Ming Zhou. 2020.
\newblock \href {http://arxiv.org/abs/2004.01401} {{XGLUE:} {A} new benchmark
  dataset for cross-lingual pre-training, understanding and generation}.
\newblock \emph{CoRR}, abs/2004.01401.

\bibitem[{Liu et~al.(2019)Liu, Ott, Goyal, Du, Joshi, Chen, Levy, Lewis,
  Zettlemoyer, and Stoyanov}]{RoBERTa}
Yinhan Liu, Myle Ott, Naman Goyal, Jingfei Du, Mandar Joshi, Danqi Chen, Omer
  Levy, Mike Lewis, Luke Zettlemoyer, and Veselin Stoyanov. 2019.
\newblock Roberta: A robustly optimized bert pretraining approach.
\newblock \emph{ArXiv}, abs/1907.11692.

\bibitem[{Malmsten et~al.(2020)Malmsten, B{\"{o}}rjeson, and
  Haffenden}]{SwedeBERT}
Martin Malmsten, Love B{\"{o}}rjeson, and Chris Haffenden. 2020.
\newblock \href {http://arxiv.org/abs/2007.01658} {Playing with words at the
  national library of sweden - making a swedish {BERT}}.
\newblock \emph{CoRR}, abs/2007.01658.

\bibitem[{Martin et~al.(2019)Martin, M{\"{u}}ller, Su{\'{a}}rez, Dupont,
  Romary, de~la Clergerie, Seddah, and Sagot}]{FrenchBERT2}
Louis Martin, Benjamin M{\"{u}}ller, Pedro Javier~Ortiz Su{\'{a}}rez, Yoann
  Dupont, Laurent Romary, {\'{E}}ric~Villemonte de~la Clergerie, Djam{\'{e}}
  Seddah, and Beno{\^{\i}}t Sagot. 2019.
\newblock \href {http://arxiv.org/abs/1911.03894} {Camembert: a tasty french
  language model}.
\newblock \emph{CoRR}, abs/1911.03894.

\bibitem[{Nguyen and Tuan~Nguyen(2020)}]{VietBERT}
Dat~Quoc Nguyen and Anh Tuan~Nguyen. 2020.
\newblock \href {https://doi.org/10.18653/v1/2020.findings-emnlp.92}
  {{P}ho{BERT}: Pre-trained language models for {V}ietnamese}.
\newblock In \emph{Findings of the Association for Computational Linguistics:
  EMNLP 2020}, pages 1037--1042, Online. Association for Computational
  Linguistics.

\bibitem[{O'Connor and Andreas(2021)}]{o2021context}
Joe O'Connor and Jacob Andreas. 2021.
\newblock What context features can transformer language models use?
\newblock In \emph{ACL/IJCNLP}.

\bibitem[{Pham et~al.(2021)Pham, Bui, Mai, and Nguyen}]{pham2020out}
Thang~M. Pham, Trung Bui, Long Mai, and Anh~M Nguyen. 2021.
\newblock Out of order: How important is the sequential order of words in a
  sentence in natural language understanding tasks?
\newblock \emph{ArXiv}, abs/2012.15180.

\bibitem[{Pires et~al.(2019)Pires, Schlinger, and Garrette}]{how_m_is_mBERT}
Telmo Pires, Eva Schlinger, and Dan Garrette. 2019.
\newblock \href {https://doi.org/10.18653/v1/P19-1493} {How multilingual is
  multilingual {BERT}?}
\newblock In \emph{Proceedings of the 57th Annual Meeting of the Association
  for Computational Linguistics}, pages 4996--5001, Florence, Italy.
  Association for Computational Linguistics.

\bibitem[{Polignano et~al.(2019)Polignano, Basile, Degemmis, Semeraro, and
  Basile}]{ItalianBERT}
Marco Polignano, Pierpaolo Basile, Marco Degemmis, Giovanni Semeraro, and
  Valerio Basile. 2019.
\newblock Alberto: Italian bert language understanding model for nlp
  challenging tasks based on tweets.
\newblock In \emph{CLiC-it}.

\bibitem[{Popovi{\'c}(2015)}]{chrf}
Maja Popovi{\'c}. 2015.
\newblock \href {https://doi.org/10.18653/v1/W15-3049} {chr{F}: character
  n-gram {F}-score for automatic {MT} evaluation}.
\newblock In \emph{Proceedings of the Tenth Workshop on Statistical Machine
  Translation}, pages 392--395, Lisbon, Portugal. Association for Computational
  Linguistics.

\bibitem[{Radford and Narasimhan(2018)}]{GPT-1}
Alec Radford and Karthik Narasimhan. 2018.
\newblock Improving language understanding by generative pre-training.

\bibitem[{Radford et~al.(2019)Radford, Wu, Child, Luan, Amodei, and
  Sutskever}]{GPT-2}
Alec Radford, Jeff Wu, Rewon Child, David Luan, Dario Amodei, and Ilya
  Sutskever. 2019.
\newblock Language models are unsupervised multitask learners.

\bibitem[{Sankar et~al.(2019)Sankar, Subramanian, Pal, Chandar, and
  Bengio}]{sankar-etal-2019-neural}
Chinnadhurai Sankar, Sandeep Subramanian, Chris Pal, Sarath Chandar, and Yoshua
  Bengio. 2019.
\newblock \href {https://doi.org/10.18653/v1/P19-1004} {Do neural dialog
  systems use the conversation history effectively? an empirical study}.
\newblock In \emph{Proceedings of the 57th Annual Meeting of the Association
  for Computational Linguistics}, pages 32--37, Florence, Italy. Association
  for Computational Linguistics.

\bibitem[{Sennrich et~al.(2015)Sennrich, Haddow, and Birch}]{BPE}
Rico Sennrich, Barry Haddow, and Alexandra Birch. 2015.
\newblock \href {http://arxiv.org/abs/1508.07909} {Neural machine translation
  of rare words with subword units}.
\newblock \emph{CoRR}, abs/1508.07909.

\bibitem[{Sinha et~al.(2021)Sinha, Jia, Hupkes, Pineau, Williams, and
  Kiela}]{sinha2021masked}
Koustuv Sinha, Robin Jia, Dieuwke Hupkes, Joelle Pineau, Adina Williams, and
  Douwe Kiela. 2021.
\newblock Masked language modeling and the distributional hypothesis: Order
  word matters pre-training for little.
\newblock \emph{arXiv preprint arXiv:2104.06644}.

\bibitem[{Sinha et~al.(2020)Sinha, Parthasarathi, Pineau, and
  Williams}]{sinha2020unnatural}
Koustuv Sinha, Prasanna Parthasarathi, Joelle Pineau, and Adina Williams. 2020.
\newblock Unnatural language inference.
\newblock \emph{arXiv preprint arXiv:2101.00010}.

\bibitem[{Srinivasan et~al.(2021)Srinivasan, Sitaram, Ganu, Dandapat, Bali, and
  Choudhury}]{PredictingPerformanceMultilingual}
Anirudh Srinivasan, Sunayana Sitaram, Tanuja Ganu, Sandipan Dandapat, Kalika
  Bali, and Monojit Choudhury. 2021.
\newblock \href {http://arxiv.org/abs/2110.08875} {Predicting the performance
  of multilingual {NLP} models}.
\newblock \emph{CoRR}, abs/2110.08875.

\bibitem[{Taktasheva et~al.(2021)Taktasheva, Mikhailov, and
  Artemova}]{ShakingTrees}
Ekaterina Taktasheva, Vladislav Mikhailov, and Ekaterina Artemova. 2021.
\newblock \href {http://arxiv.org/abs/2109.14017} {Shaking syntactic trees on
  the sesame street: Multilingual probing with controllable perturbations}.
\newblock \emph{CoRR}, abs/2109.14017.

\bibitem[{Wang et~al.(2019{\natexlab{a}})Wang, Pruksachatkun, Nangia, Singh,
  Michael, Hill, Levy, and Bowman}]{SuperGLUE}
Alex Wang, Yada Pruksachatkun, Nikita Nangia, Amanpreet Singh, Julian Michael,
  Felix Hill, Omer Levy, and Samuel~R. Bowman. 2019{\natexlab{a}}.
\newblock \href {http://arxiv.org/abs/1905.00537} {Superglue: {A} stickier
  benchmark for general-purpose language understanding systems}.
\newblock \emph{CoRR}, abs/1905.00537.

\bibitem[{Wang et~al.(2019{\natexlab{b}})Wang, Singh, Michael, Hill, Levy, and
  Bowman}]{GLUE}
Alex Wang, Amanpreet Singh, Julian Michael, Felix Hill, Omer Levy, and
  Samuel~R. Bowman. 2019{\natexlab{b}}.
\newblock {GLUE:} {A} multi-task benchmark and analysis platform for natural
  language understanding.
\newblock In \emph{7th International Conference on Learning Representations,
  {ICLR} 2019, New Orleans, LA, USA, May 6-9, 2019}.

\bibitem[{Wu et~al.(2016)Wu, Schuster, Chen, Le, Norouzi, Macherey, Krikun,
  Cao, Gao, Macherey, Klingner, Shah, Johnson, Liu, Kaiser, Gouws, Kato, Kudo,
  Kazawa, Stevens, Kurian, Patil, Wang, Young, Smith, Riesa, Rudnick, Vinyals,
  Corrado, Hughes, and Dean}]{wordpiece}
Yonghui Wu, Mike Schuster, Zhifeng Chen, Quoc~V. Le, Mohammad Norouzi, Wolfgang
  Macherey, Maxim Krikun, Yuan Cao, Qin Gao, Klaus Macherey, Jeff Klingner,
  Apurva Shah, Melvin Johnson, Xiaobing Liu, Lukasz Kaiser, Stephan Gouws,
  Yoshikiyo Kato, Taku Kudo, Hideto Kazawa, Keith Stevens, George Kurian,
  Nishant Patil, Wei Wang, Cliff Young, Jason Smith, Jason Riesa, Alex Rudnick,
  Oriol Vinyals, Greg Corrado, Macduff Hughes, and Jeffrey Dean. 2016.
\newblock \href {http://arxiv.org/abs/1609.08144} {Google's neural machine
  translation system: Bridging the gap between human and machine translation}.
\newblock \emph{CoRR}, abs/1609.08144.

\bibitem[{Xia et~al.(2020)Xia, Anastasopoulos, Xu, Yang, and
  Neubig}]{PredictingPerformanceNLP}
Mengzhou Xia, Antonios Anastasopoulos, Ruochen Xu, Yiming Yang, and Graham
  Neubig. 2020.
\newblock \href {https://doi.org/10.18653/v1/2020.acl-main.764} {Predicting
  performance for natural language processing tasks}.
\newblock In \emph{Proceedings of the 58th Annual Meeting of the Association
  for Computational Linguistics}, pages 8625--8646, Online. Association for
  Computational Linguistics.

\bibitem[{Ye et~al.(2021)Ye, Liu, Fu, and
  Neubig}]{PredictingPerformanceNLPFinegrain}
Zihuiwen Ye, Pengfei Liu, Jinlan Fu, and Graham Neubig. 2021.
\newblock \href {https://doi.org/10.18653/v1/2021.eacl-main.324} {Towards more
  fine-grained and reliable {NLP} performance prediction}.
\newblock In \emph{Proceedings of the 16th Conference of the European Chapter
  of the Association for Computational Linguistics: Main Volume}, pages
  3703--3714, Online. Association for Computational Linguistics.

\end{thebibliography}
\bibliographystyle{acl_natbib}

\appendix

\onecolumn
\section{Experiment Details}\label{app:experiments}

\paragraph{Model Hyperparameters and Training}\label{app:hyperparameters_training}

We finetune each pretrained models on the English version of each dataset for a total of 10 epochs, checkpointing the model after each epochs.
The English version is never perturbed, the finetuning is done on unperturbed data.
This finetuning is done 5 times with different random seeds for each model and each datasets.
For 7 datasets and 3 models we have a total of $3 * 7 * 5 = 105$ finetuning and $1050$ checkpoints, one for each epoch.
A learning rate of 2e-5, a batch size of 32 and a weight decay of 0.01 is used in all finetuning.
All experiments used a warmup ratio of 0.06, as described in \citet{RoBERTa}.
A maximum sequence length of 512 for the mBERT and XLM-R model and a maximum sequence length of 2048 for the Canine model are used.

For the evaluation, we perform the same perturbations on the validation and testing data of the different target languages.
We evaluate the perturbed validation data on each of the 10 checkpoints, chose the best checkpoint on the perturbed validation data, and evaluate that checkpoint on the perturbed test data.
This process is repeated for each perturbations, each of the 5 random seed and 5 times with different perturbation random seeds for each finetuned models.
In total, for each language in each task on each model for each perturbation setup we average results over 25 random seeds.

For the sentence retrieval tasks, such as Tatoeba and BUCC, we do not perform any finetuning.
We obtain the representation by averaging the output of the final hidden layer of the model.~\citep{XTREME}
First, we obtain the representation of the unperturbed English side of the dataset.
This is done by feeding the English text through the model and averaging the final layers hidden representation of the text.
We then perform our perturbations on the target language text, feed those perturbed text through the same pretrained cross-lingual model and obtain it's representation through the same process.
We now have a set of English representation and a set of target language representation, on which we can obtain the cosine distances.
We can either find the nearest neighbours (Tatoeba, BUCC) or use the Z-Score of those representations (MTData).
If the nearest neighbour is the sentence that was to be retrieved, we consider this an hit, else it is a miss.
The reported results are over the average of 5 random seeds of those perturbations.

\paragraph{Monolingual Local Sensitivity}\label{app:monolingual}
The monolingual sentence retrieval task is performed in the exact same process as for the sentence retrieval task described in Appendix \ref{app:hyperparameters_training}.
The only difference is that the unperturbed English text is replaced by the target language corpus.
Pictured in Figure~\ref{fig:monolingual_perturbations} is a toy example representing the monolingual sentence retrieval tasks compared to the crosslingual one.
We calculate monolingual local sensitivity by taking the correlation of the degradation in performance on the monolingual sentence retrieval task with the amount of local perturbations applied to the right side of the dataset.

\begin{figure}[ht]
    \centering
    \includegraphics[width=0.95\textwidth]{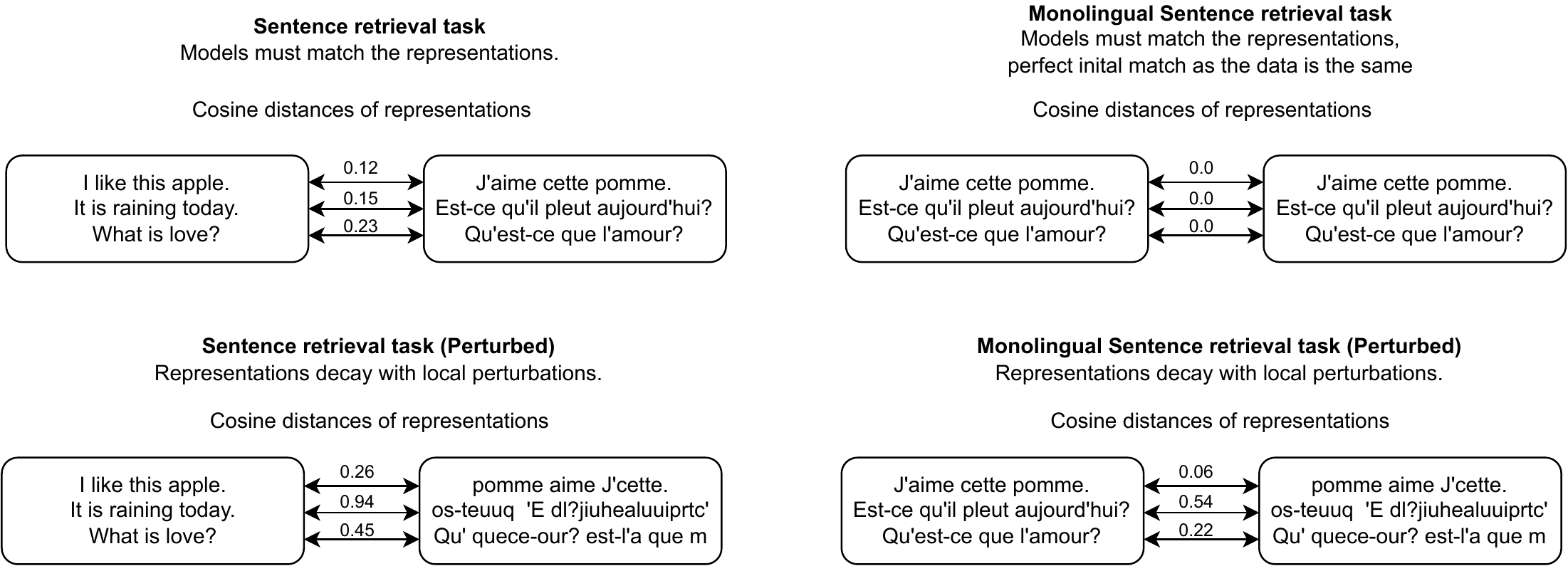}
    \caption{Toy example of sentence retrieval and monolingual sentence retrieval with and without perturbations.}
    \label{fig:monolingual_perturbations}
\end{figure}

\paragraph{Perturbations}\label{app:perturbations}
A total of 13 evaluations, containing 12 perturbations are used for all experiments.
The first one is the Benchmark, which is simply the unperturbed text.
On a character-level perturbations we perform neighbour-flip shuffling with $\rho$ values of: $[0.025$, $0.05$, $0.075$, $0.1$, $0.125$, $0.15$, $0.175$, $0.2$, $0.25$, $0.3$, $0.35$, $0.45]$.
No neighbor-flip with $\rho$ over $0.5$ or over are performed, as they would ultimately shuffle the text \textit{less}.
Unlike \citet{clouatre-etal-2022-local}, we focus purely on local structure perturbations, as we are not interested in the relative importance of local structure compared to other structures, but simply that local structure is important at all.

\newpage
\section{Additional Results}

\paragraph{Cross-Lingual Local Sensitivity Additional Results}\label{app:results_cross_lingual}
In this section we present additional results on the first set of experiments on the cross-lingual zero-shot local sensitivity tasks.

\begin{figure}[ht]
    \centering
    \includegraphics[width=0.98\columnwidth]{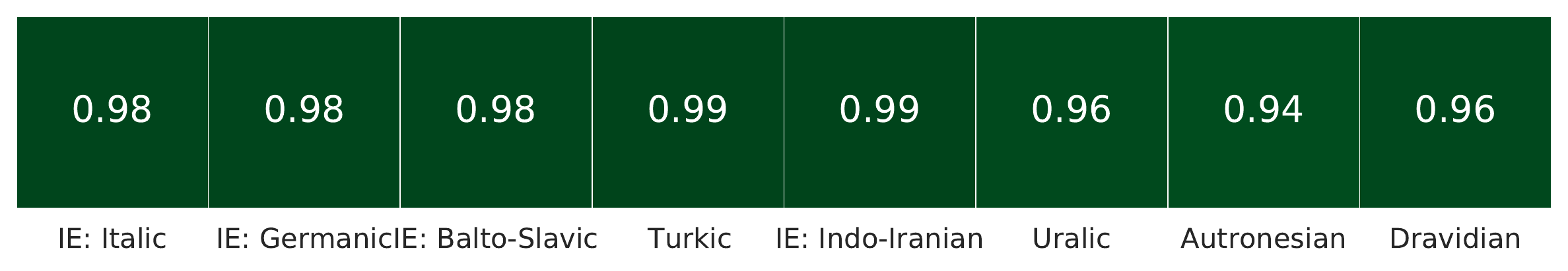}
    \caption{Local sensitivity matrix between the different languages families with at least 3 tested languages in our tasks, averaged across all tasks and models.}
    \label{fig:family-metric-correlation}
\end{figure}

\begin{figure}[ht]
    \centering
    \includegraphics[width=0.98\columnwidth]{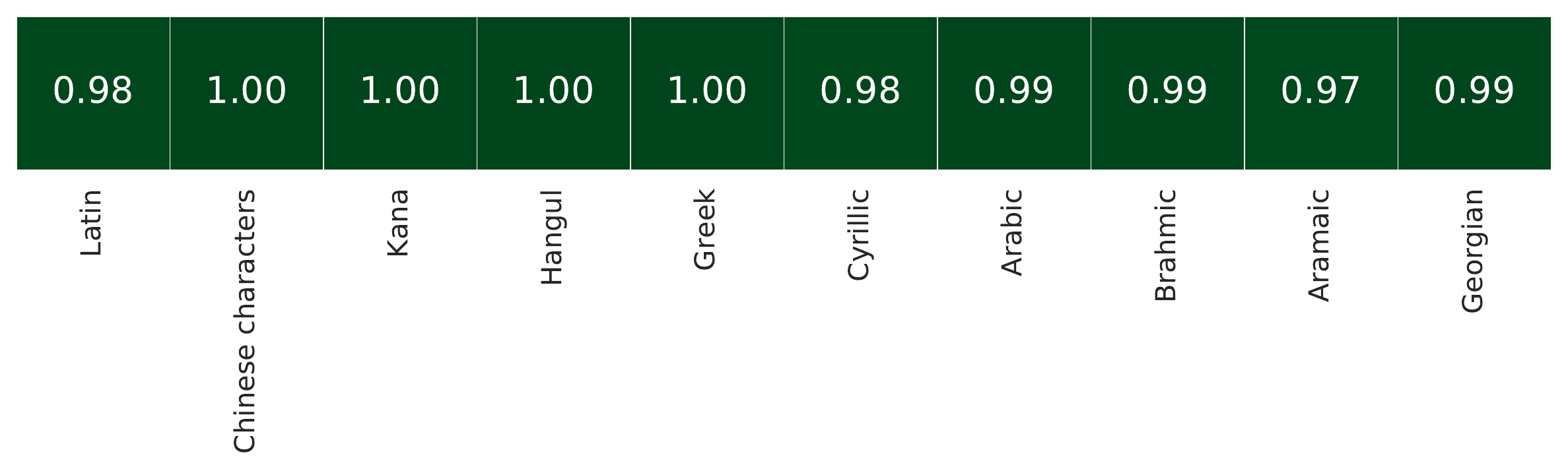}
    \caption{Local sensitivity matrix between the different scripts with at least 3 tested languages in our tasks, averaged across all tasks and models.}
    \label{fig:script-metric-correlation}
\end{figure}

The trend of extremely high correlation between performance and perturbations also holds when grouping results by script and language family, as shown in Figure~\ref{fig:family-metric-correlation} and Figure~\ref{fig:script-metric-correlation}.

\begin{figure}[ht]
    \centering
    \subfigure[Tatoeba, QAM, WPR]{
    \includegraphics[width=0.32\textwidth]{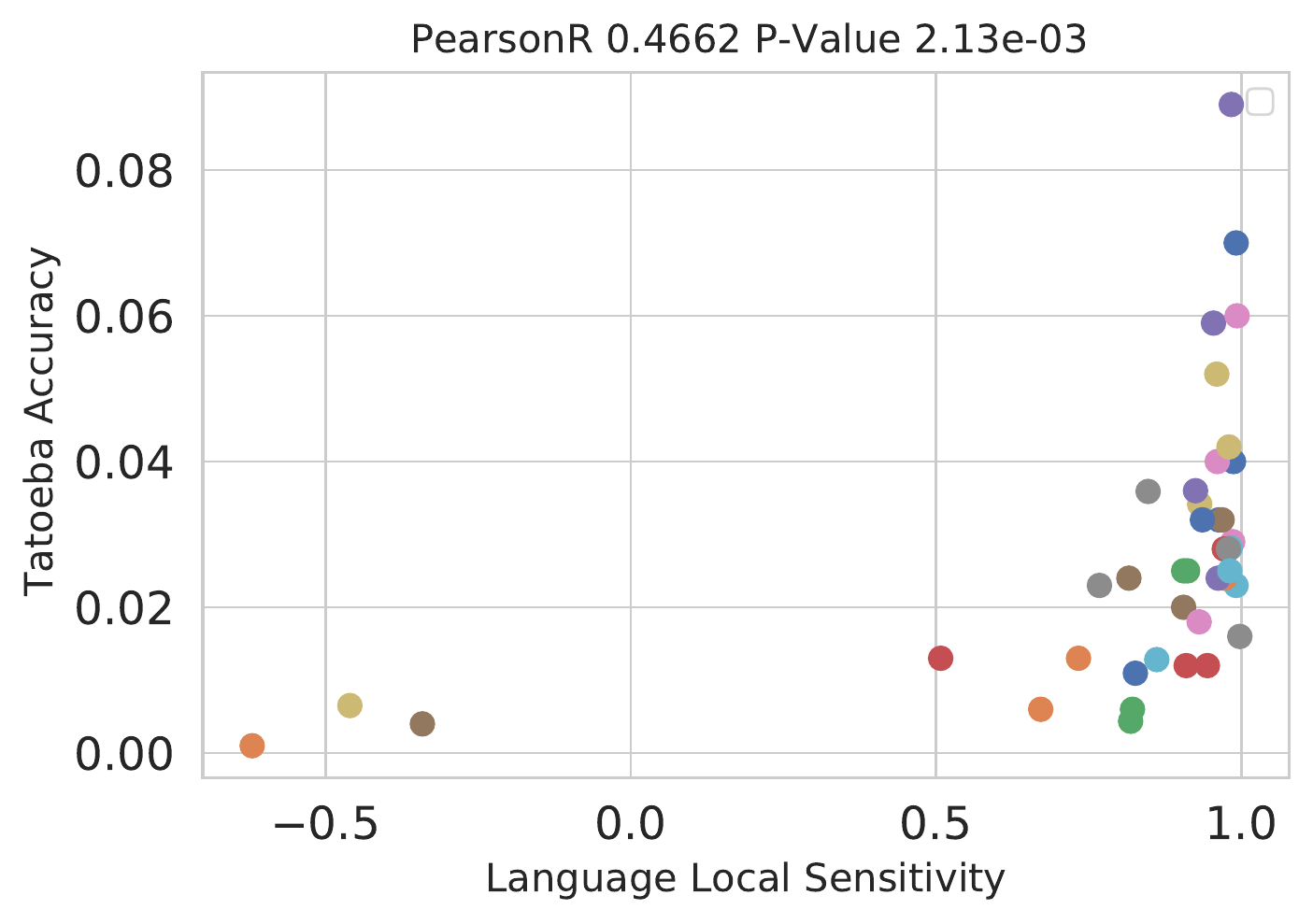}\hfill
    \includegraphics[width=0.32\textwidth]{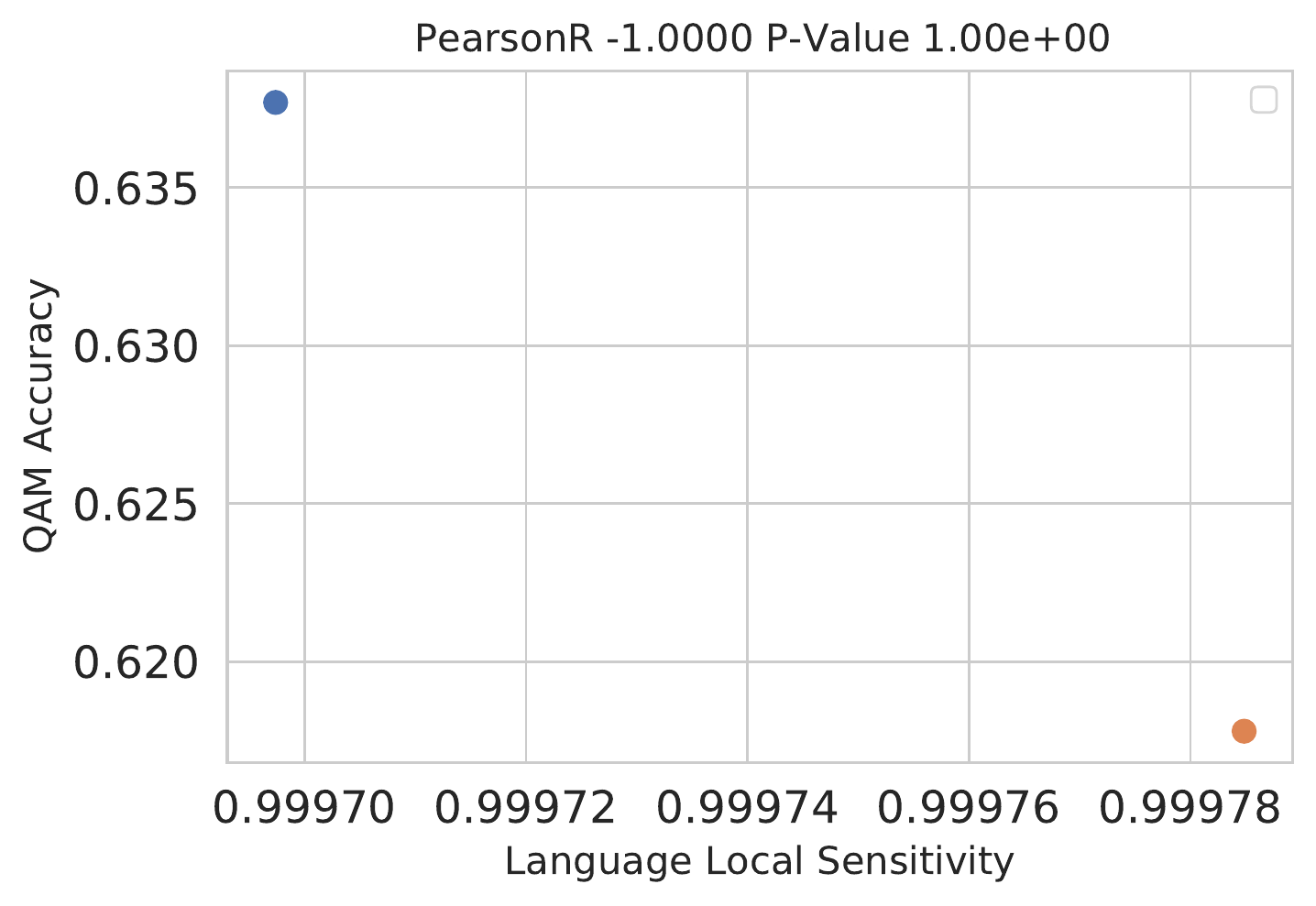}\hfill
    \includegraphics[width=0.32\textwidth]{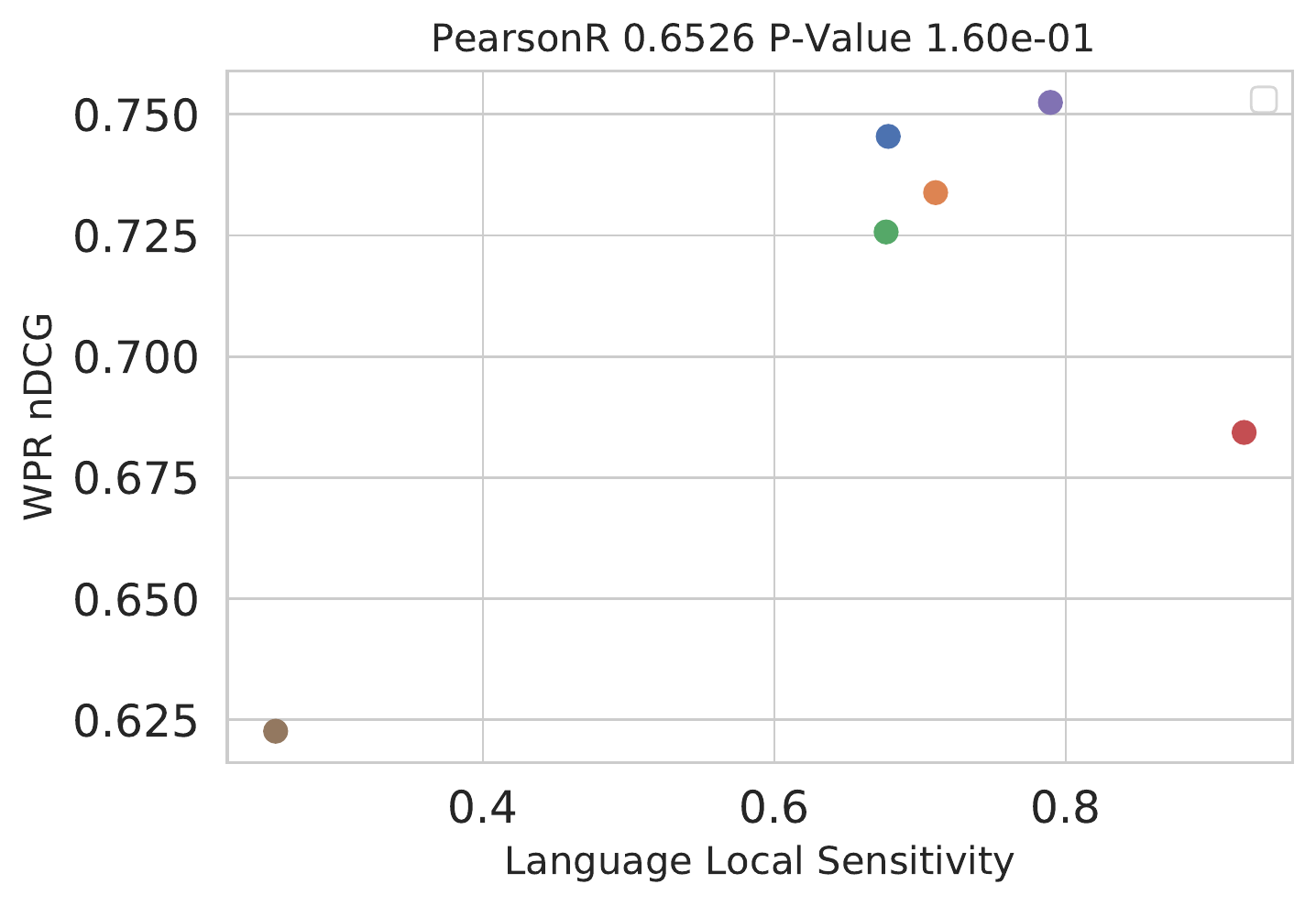}}
   
    \subfigure[Paws-X, BUCC, QADSM]{
    \includegraphics[width=0.32\textwidth]{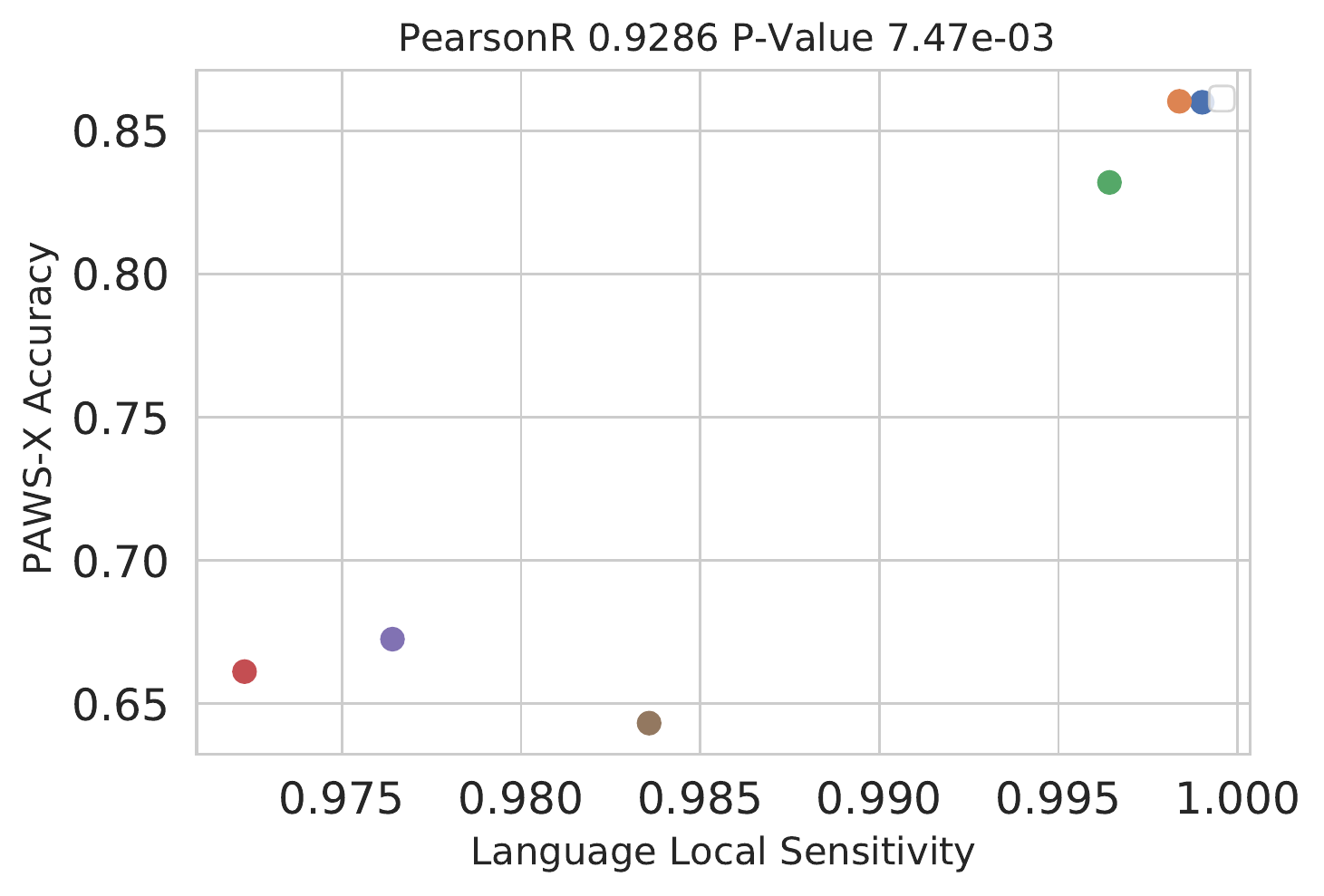}\hfill
    \includegraphics[width=0.32\textwidth]{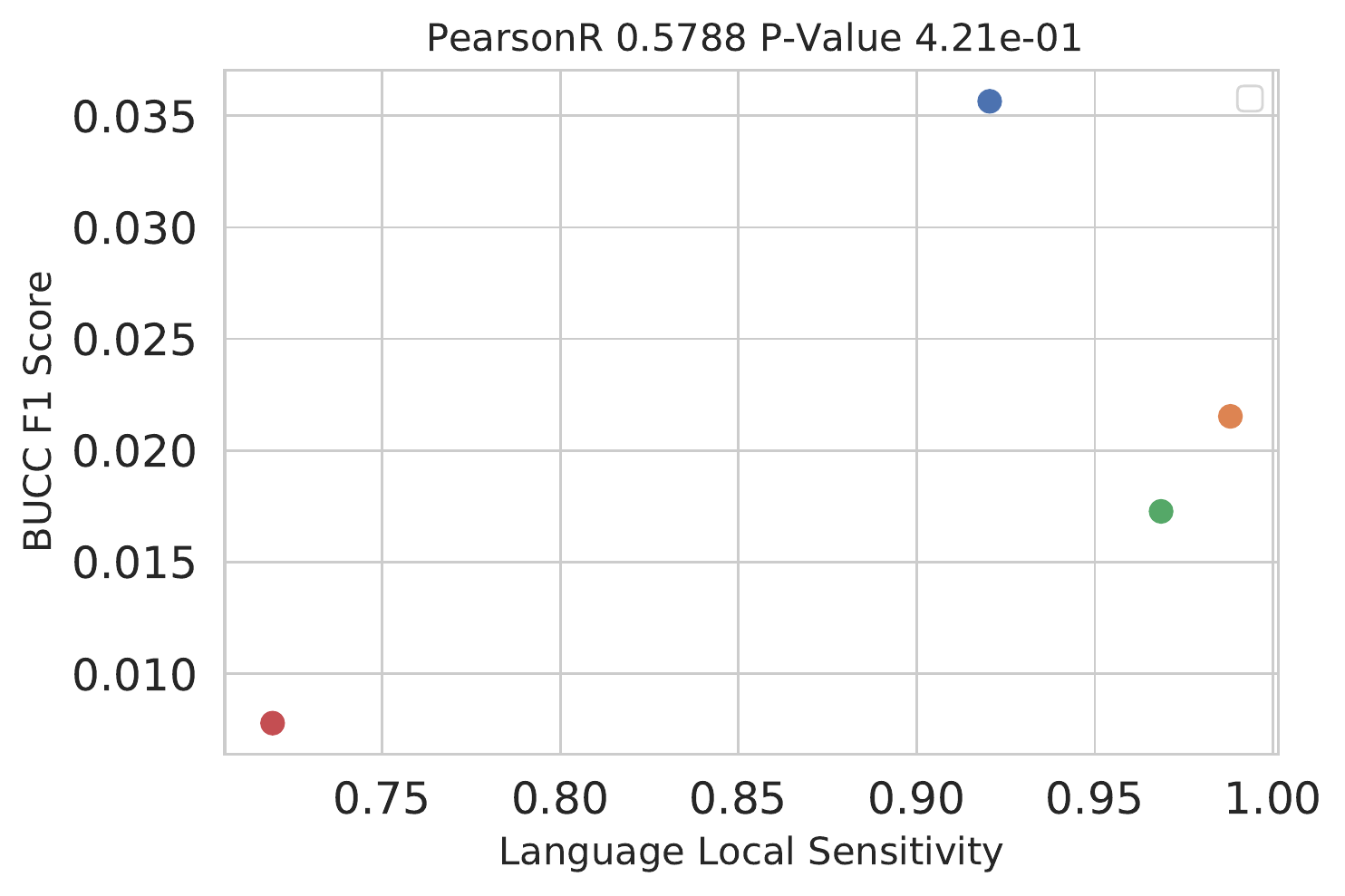}\hfill
    \includegraphics[width=0.32\textwidth]{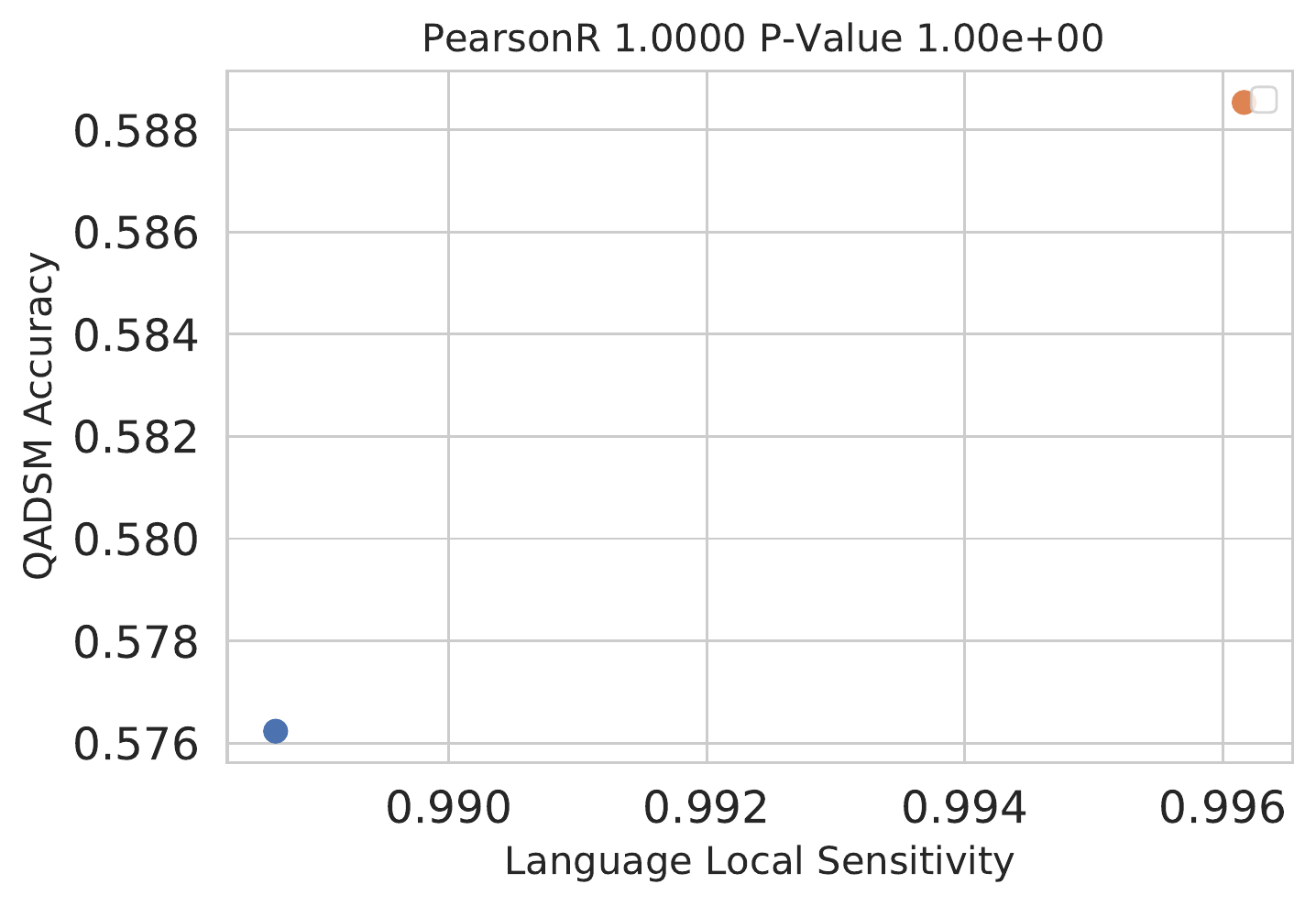}}

    \caption{Plotted are the individual language's local sensitivity plotted against their performance on the unperturbed text on all tasks, averaged across all models.
    We can observe that with the exception of QAM, which only contains two language with very high local sensitivity, all language and tasks exhibit the same overall behaviour.
    Languages with low local sensitivity invariably have low performance.
    }
    \label{fig:no_emb_metrics}
\end{figure}

Further, using low local sensitivity to predict low performance on a particular language seem to be consistent across tested tasks, as seen in Figure~\ref{fig:no_emb_metrics}. 

\newpage
\paragraph{Low-Performance Languages}\label{app:low_ressource}
\begin{figure}[ht]
    \centering
    \includegraphics[width=0.98\textwidth]{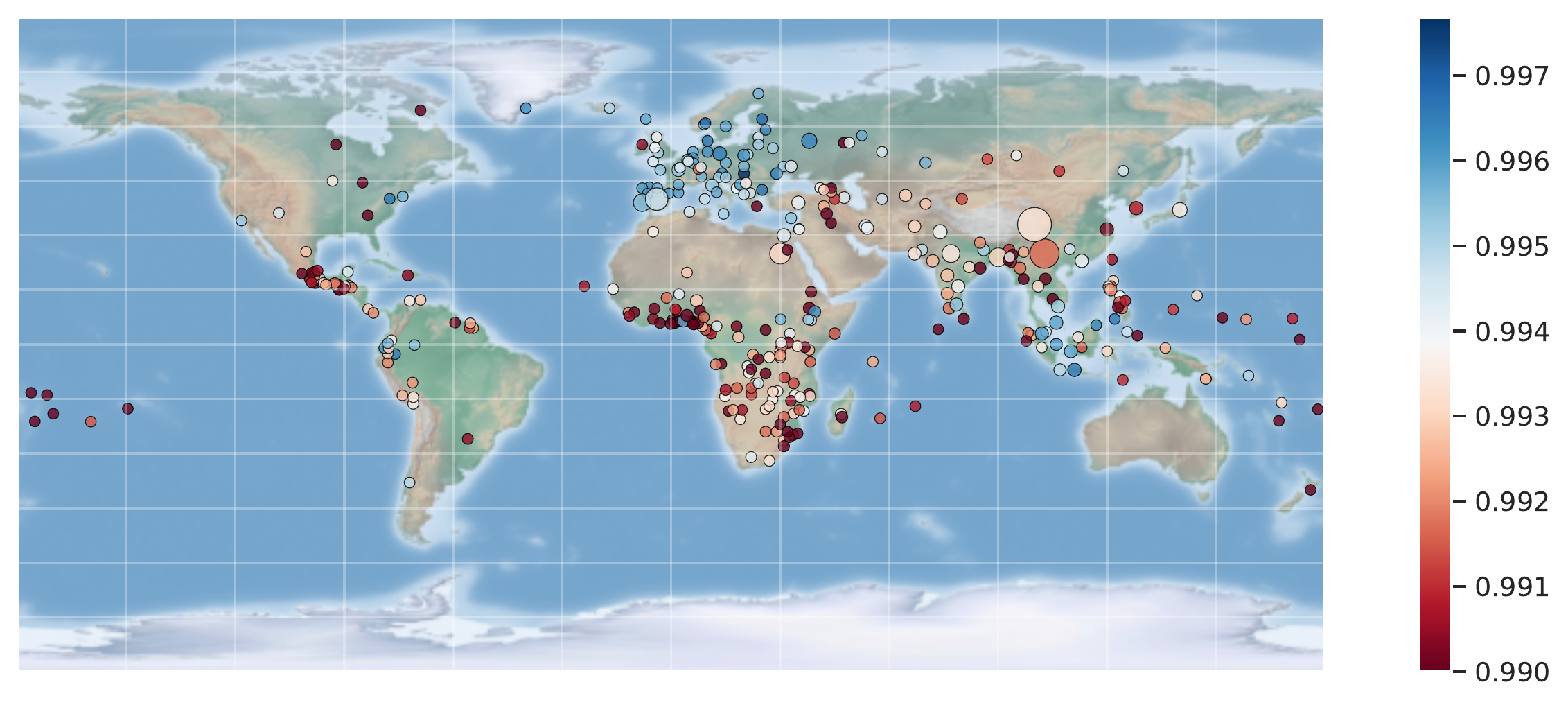}\hfill

    \caption{
    Monolingual local sensitivity of $350$ languages on the task of cross-lingual similarity on our MTData cross-lingual sentence similarity dataset, scaled by the estimated amount of native speakers.}
    \label{fig:world_map_correlations}
\end{figure}
\begin{table}[ht]
\centering
\begin{adjustbox}{width=0.95\textwidth}
\small
\begin{tabular}{||c c c c c c c c c||} 
 \hline
 \textbf{Languages} & Coptic & Northwestern Ojibwa & Inuktitut & Lao & Dhivehi & S'gaw Karen & Yoruba & Khmer\\ [0.5ex] 
 \hline\hline
 \textbf{Sensitivity} & $0.973$ & $0.978$ & $0.979$ & $0.979$ & $0.980$ & $0.981$ & $0.982$ & $0.73$\\ 
 \hline
 \textbf{Family} & Afro-Asiatic & Algic & Eskimo–Aleut & Kra-Dai & IE: Indo-Iranian & Sino-Tibetan & Niger-Congo & Austroasiatic\\
 \hline
 \textbf{Script} & Coptic & Latin & Inuktitut syllabics & Brahmic &  Thaana & Brahmic & Latin & Brahmic\\ 
  \hline
 \textbf{Native Speaker (Millions)} & $0.0$ & $0.02$ & $0.04$ & $30$ & $0.34$ & $3$ & $43$ & $16$\\
 \hline\hline
 
 \textbf{Languages} & Maori & Sinhala & Samoan & Cherokee & Syriac & Nzima & Oriya & Venda\\ [0.5ex] 
 \hline\hline
 \textbf{Sensitivity} & $0.983$ & $0.984$ & $0.984$ & $0.985$ & $0.985$ & $0.985$ & $0.987$ & $0.987$\\ 
 \hline
 \textbf{Family} & Austronesian & IE: Indo-Iranian & Autronesian & Iroquoian & Afro-Asiatic & Niger-Congo & IE: Indo-Iranian & Niger-Congo\\
 \hline
 \textbf{Script} & Latin & Brahmic & Latin & Latin & Aramaic & Latin & Brahmic & Latin\\ 
  \hline
 \textbf{Native Speaker (Millions)}~\footnote{Properly cite: Ethnologue : languages of the world 2019 } & $0.05$ & $17$ & $0.51$ & $0.002$ & $0.24$ & $0.41$ & $35$ & $1.3$\\ [1ex] 
 \hline
\end{tabular}

\end{adjustbox}
\caption{Statistics on the language containing the lowest monolingual local sensitivity of all 350 languages.} 
\label{tab:weak_language_stat}
\end{table}

In Figure~\ref{fig:world_map_correlations}, we have plotted the monolingual local sensitivity of all $350$ languages on a world map at their geographical centers~\citep{wals1, wals2}, with their size scaled by amount of native speakers in those specific languages.
Many statements can be made about low-performance languages from this study.

It seems that languages that are geographically close to Europe or South-East Asia are generally well understood by our cross-lingual models.
The majority of poorly understood languages seem to either be concentrated in Sub-Saharan Africa or Central America, as well as island-specific languages across the Pacific ocean.

The languages that have the lowest monolingual local sensitivity are reported in Table~\ref{tab:weak_language_stat}.
Some of those languages, like Coptic, a now long-dead language in an unseen script, are fairly obvious low-performers.
Our approach, however, seems able to detect low-performance in languages that would not be that obvious and would be quite important to detect, like Lao with its over 30 million native speakers.

\newpage
\section{MTData Sentence Similarity Task}\label{app:mtdata}
From the MTData dataset~\citep{MTData} we build a sentence similarity dataset covering a total of 350 languages.
We use and adapt the approach used to build the Tatoeba sentence retrieval dataset~\citep{tatoeba}.
Specifically, from the MTData dataset containing english aligned sentences in over 500 languages, we remove all sentences containing either "@", "http" or "\%", remove any English sentence containing less than 3 words, and remove any duplicate.
We randomly sample 1000 sentence pairs per language, removing languages with less then 1000 sentence pair present after filtering.
We also remove text of sign languages, as their written form is almost exactly the same as the original language.
In total, 350 languages remain after that point.
Table~\ref{tab:all_mtdata_languages_1} to Table~\ref{tab:all_mtdata_languages_6} contains statistics on every single language present in our MTData Sentence Retrieval Task.

\begin{table}[ht]
\centering
\begin{adjustbox}{width=0.95\textwidth}
\small
\begin{tabular}{||c c c c c||} 
 \hline
 \textbf{Language} & \textbf{ISO-639-3} & \textbf{Language Family} & \textbf{Language Script} & \textbf{Num Native Speakers}\\ [0.5ex] 
 \hline\hline
 \hline
 Chinese & zho & Sino-Tibetan & Chinese characters and derivatives & 1300000000\\
\hline
 Mandarin Chinese & cmn & Sino-Tibetan & Chinese characters and derivatives & 920000000\\
\hline
 Spanish & spa & IE: Italic & Latin & 493000000\\
\hline
 Arabic & ara & Afro-Asiatic & Arabic & 400000000\\
\hline
 Bengali & ben & IE: Indo-Iranian & Brahmic & 300000000\\
\hline
 Hindi & hin & IE: Indo-Iranian & Brahmic & 260000000\\
\hline
 Portuguese & por & IE: Italic & Latin & 250000000\\
\hline
 Russian & rus & IE: Balto-Slavic & Cyrillic & 150000000\\
\hline
 Japanese & jpn & Japonic & Kana & 128000000\\
\hline
 Panjabi & pan & IE: Indo-Iranian & Arabic & 113000000\\
\hline
 German & deu & IE: Germanic & Latin & 95000000\\
\hline
 Yue Chinese & yue & Sino-Tibetan & Chinese characters and derivatives & 84000000\\
\hline
 Egyptian Arabic & arz & Afro-Asiatic & Arabic & 83000000\\
\hline
 Javanese & jav & Autronesian & Brahmic & 82000000\\
\hline
 Korean & kor & Koreanic & Hangul & 80400000\\
\hline
 Turkish & tur & Turkic & Latin & 80000000\\
\hline
 Wu Chinese & wuu & Sino-Tibetan & Chinese characters and derivatives & 80000000\\
\hline
 Malay (individual language) & zlm & Autronesian & Arabic & 77000000\\
\hline
 Malay (macrolanguage) & msa & Austronesian & Latin & 77000000\\
\hline
 Standard Malay & zsm & Autronesian & Arabic & 77000000\\
\hline
 French & fra & IE: Italic & Latin & 76800000\\
\hline
 Vietnamese & vie & Austroasiatic & Latin & 76000000\\
\hline
 Telugu & tel & Dravidian & Brahmic & 75000000\\
\hline
 Marathi & mar & IE: Indo-Iranian & Brahmic & 73000000\\
\hline
 Persian & fas & IE: Indo-Iranian & Arabic & 70000000\\
\hline
 Tamil & tam & Dravidian & Brahmic & 70000000\\
\hline
 Urdu & urd & IE: Indo-Iranian & Arabic & 70000000\\
\hline
 Italian & ita & IE: Italic & Latin & 67000000\\
\hline
 Iranian Persian & pes & IE: Indo-Iranian & Arabic & 55600000\\
\hline
 Gujarati & guj & IE: Indo-Iranian & Brahmic & 50000000\\
\hline
 Hausa & hau & Afro-Asiatic & Latin & 50000000\\
\hline
 Pushto & pus & IE: Indo-Iranian & Arabic & 50000000\\
\hline
 Tagalog & tgl & Autronesian & Latin & 45000000\\
\hline
 Polish & pol & IE: Balto-Slavic & Latin & 45000000\\
\hline
 Filipino & fil & Austronesian & Latin & 45000000\\
\hline
 Uzbek & uzb & Turkic & Latin & 44000000\\
\hline
 Indonesian & ind & Autronesian & Latin & 43000000\\
\hline
 Yoruba & yor & Niger-Congo & Latin & 43000000\\
\hline
 Kannada & kan & Dravidian & Brahmic & 43000000\\
\hline
 Sundanese & sun & Austronesian & Latin & 42000000\\
\hline
 Ukrainian & ukr & IE: Balto-Slavic & Cyrillic & 40000000\\
\hline
 Nigerian Pidgin & pcm & English Creole & Latin & 40000000\\
\hline
 Oromo & orm & Afro-Asiatic & Latin & 37400000\\
\hline
 Oriya (macrolanguage) & ori & IE: Indo-Iranian & Brahmic & 35000000\\
\hline
 Malayalam & mal & Dravidian & Brahmic & 35000000\\
\hline
 Maithili & mai & IE: Indo-Iranian & Brahmic & 33900000\\
\hline
 Burmese & mya & Sino-Tibetan & Brahmic & 33000000\\
\hline
 Amharic & amh & Afro-Asiatic & Ge'ez & 32000000\\
\hline
 Azerbaijani & aze & Turkic & Arabic & 30000000\\
\hline
 Lao & lao & Kra-Dai & Brahmic & 30000000\\
\hline
 Igbo & ibo & Niger-Congo & Latin & 30000000\\
\hline
 Thai & tha & Kra-Dai & Brahmic & 28000000\\
\hline
 Sindhi & snd & IE: Indo-Iranian & Arabic & 25000000\\
\hline
 Malagasy & mlg & Austronesian & Latin & 25000000\\
\hline
 Plateau Malagasy & plt & Austronesian & Latin & 25000000\\
\hline
 Dutch & nld & IE: Germanic & Latin & 25000000\\
\hline
 Kurdish & kur & IE: Indo-Iranian & Arabic & 25000000\\
\hline
 Romanian & ron & IE: Italic & Latin & 23800000\\
\hline
 Cebuano & ceb & Autronesian & Latin & 22000000\\
\hline
 Somali & som & Afro-Asiatic & Latin & 21807730\\
\hline
 Croatian & hrv & IE: Balto-Slavic & Cyrillic & 21000000\\
\hline
 Ganda & lug & Niger-Congo & Latin & 20000000\\
\hline
 Ewe & ewe & Niger-Congo & Latin & 20000000\\
\hline
 Swahili (macrolanguage) & swa & Niger-Congo & Latin & 18000000\\
\hline
 Chhattisgarhi & hne & IE: Indo-Iranian & Brahmic & 18000000\\
\hline
 Kazakh & kaz & Turkic & Cyrillic & 17800000\\
\hline
 Lingala & lin & Niger-Congo & Latin & 17500000\\
\hline
 Sinhala & sin & IE: Indo-Iranian & Brahmic & 17000000\\
\hline
 Nepali (macrolanguage) & nep & IE: Indo-Iranian & Brahmic & 16000000\\
 
 \hline
\end{tabular}

\end{adjustbox}
\caption{Statistics on all 350 languages present in the MTData sentence retrieval dataset. (1 of 6)} 
\label{tab:all_mtdata_languages_1}
\end{table}

\begin{table}[ht]
\centering
\begin{adjustbox}{width=0.95\textwidth}
\small
\begin{tabular}{||c c c c c||} 
 \hline
 \textbf{Language} & \textbf{ISO-639-3} & \textbf{Language Family} & \textbf{Language Script} & \textbf{Num Native Speakers}\\ [0.5ex] 
 \hline\hline
\hline
 Khmer & khm & Austroasiatic & Brahmic & 16000000\\
\hline
 Assamese & asm & IE: Indo-Iranian & Brahmic & 15311351\\

\hline
 Northern Kurdish & kmr & IE: Indo-Iranian & Arabic & 15000000\\
\hline
 Bavarian & bar & IE: Germanic & Latin & 14000000\\
\hline
 Modern Greek (1453-) & ell & IE: Hellenic & Greek & 13400000\\
\hline
 Hungarian & hun & Uralic & Latin & 13000000\\
\hline
 Umbundu & umb & Niger-Congo & Latin & 12740000\\
\hline
 Haitian & hat & IE: Italic & Latin & 12000000\\
\hline
 Shona & sna & Niger-Congo & Latin & 12000000\\
\hline
 Zulu & zul & Niger-Congo & Latin & 12000000\\
\hline
 Serbian & srp & IE: Balto-Slavic & Cyrillic & 12000000\\
\hline
 Nyanja & nya & Niger-Congo & Latin & 12000000\\
\hline
 Rundi & run & Niger-Congo & Latin & 11244750\\
\hline
 Turkmen & tuk & Turkic & Latin & 11000000\\
\hline
 Czech & ces & IE: Balto-Slavic & Latin & 10700000\\
\hline
 Swedish & swe & IE: Germanic & Latin & 10000000\\
\hline
 Uighur & uig & Turkic & Arabic & 10000000\\
\hline
 Tigrinya & tir & Afro-Asiatic & Ge'ez & 9850000\\

 Kinyarwanda & kin & Niger-Congo & Latin & 9800000\\
\hline
 Congo Swahili & swc & Niger-Congo & Latin & 9000000\\
\hline
 Xhosa & xho & Niger-Congo & Latin & 8700000\\
\hline
 Ga & gaa & Niger-Congo & Latin & 8500000\\
\hline
 Iloko & ilo & Austronesian & Latin & 8100000\\
\hline
 Tajik & tgk & IE: Indo-Iranian & Cyrillic & 8100000\\
\hline
 Bulgarian & bul & IE: Balto-Slavic & Cyrillic & 8000000\\
\hline
 Quechua & que & Quechuan & Latin & 8000000\\
\hline
 Mossi & mos & Niger-Congo & Latin & 7830000\\
\hline
 Hiligaynon & hil & Austronesian & Latin & 7800000\\
\hline
 Makhuwa & vmw & Niger-Congo & Latin & 7400000\\
\hline
 Afrikaans & afr & IE: Germanic & Arabic & 7200000\\
\hline
 Dyula & dyu & Mande & Latin & 6852620\\
\hline
 Kikuyu & kik & Niger-Congo & Latin & 6600000\\
\hline
 Paraguayan Guaraní & gug & Tupian & Latin & 6500000\\
\hline
 San Salvador Kongo & kwy & Niger-Congo & Latin & 6500000\\
\hline
 Kongo & kon & Niger-Congo & Latin & 6500000\\
\hline
 Luba-Lulua & lua & Niger-Congo & Latin & 6300000\\
\hline
 Low German & nds & IE: Germanic & Latin & 6000000\\
\hline
 Armenian & hye & IE: Armenian & Armenian & 6000000\\
\hline
 Albanian & sqi & IE: Albanian & Latin & 6000000\\
\hline
 Danish & dan & IE: Germanic & Latin & 6000000\\
\hline
 Kabyle & kab & Afro-Asiatic & Arabic & 6000000\\
\hline
 Finnish & fin & Uralic & Latin & 5800000\\
\hline
 Wolof & wol & Niger-Congo & Latin & 5454000\\
\hline
 Norwegian & nor & IE: Germanic & Latin & 5320000\\
\hline
 Slovak & slk & IE: Balto-Slavic & Latin & 5200000\\
\hline
 Tatar & tat & Turkic & Cyrillic & 5200000\\
\hline
 Tswana & tsn & Niger-Congo & Latin & 5200000\\
\hline
 Mongolian & mon & Mongolic & Cyrillic & 5200000\\
\hline
 Belarusian & bel & IE: Balto-Slavic & Cyrillic & 5100000\\
\hline
 Tiv & tiv & Niger-Congo & Latin & 5000000\\
\hline
 Hebrew & heb & Afro-Asiatic & Aramaic & 5000000\\
\hline
 Pedi & nso & Niger-Congo & Latin & 4700000\\
\hline
 Baoulé & bci & Niger-Congo & Latin & 4700000\\
\hline
 Kirghiz & kir & Turkic & Cyrillic & 4500000\\
\hline
 Luo (Kenya and Tanzania) & luo & Nilo-Saharan & Latin & 4200000\\
\hline
 Bemba (Zambia) & bem & Niger-Congo & Latin & 4100000\\
\hline
 Kamba (Kenya) & kam & Niger-Congo & Latin & 3900000\\
\hline
 Tachelhit & shi & Afro-Asiatic & Arabic & 3900000\\
  
 \hline
\end{tabular}

\end{adjustbox}
\caption{Statistics on all 350 languages present in the MTData sentence retrieval dataset. (2 of 6)} 
\label{tab:all_mtdata_languages_2}
\end{table}

\begin{table}[ht]
\centering
\begin{adjustbox}{width=0.95\textwidth}
\small
\begin{tabular}{||c c c c c||} 
 \hline
 \textbf{Language} & \textbf{ISO-639-3} & \textbf{Language Family} & \textbf{Language Script} & \textbf{Num Native Speakers}\\ [0.5ex] 
 \hline\hline
 
\hline
 Lombard & lmo & IE: Italic & Latin & 3800000\\
\hline
 Georgian & kat & Kartvelian & Georgian & 3700000\\
\hline
 Hmong & hmn & Hmong–Mien & Latin & 3700000\\
\hline
 Tsonga & tso & Niger-Congo & Latin & 3700000\\
\hline
 Waray (Philippines) & war & Autronesian & Latin & 3600000\\

\hline
 Zarma & dje & Nilo-Saharan & Latin & 3600000\\
\hline
 Tumbuka & tum & Niger-Congo & Latin & 3546000\\
\hline
 Romany & rom & IE: Indo-Iranian & Latin & 3500000\\
\hline
 Nyankole & nyn & Niger-Congo & Latin & 3400000\\
\hline
 Yao & yao & Niger-Congo & Latin & 3100000\\
\hline
 Lithuanian & lit & IE: Balto-Slavic & Latin & 3000000\\
\hline
 S'gaw Karen & ksw & Sino-Tibetan & Brahmic & 3000000\\
\hline
 Sidamo & sid & Afro-Asiatic & Latin & 3000000\\
\hline
 Pampanga & pam & Autronesian & Brahmic & 2800000\\
\hline
 Slovenian & slv & IE: Balto-Slavic & Latin & 2500000\\
\hline
 Macedonian & mkd & IE: Balto-Slavic & Cyrillic & 2500000\\
\hline
 Bosnian & bos & IE: Balto-Slavic & Cyrillic & 2500000\\
\hline
 Central Bikol & bcl & Austronesian & Latin & 2500000\\
\hline
 Galician & glg & IE: Italic & Latin & 2400000\\
\hline
 Ndau & ndc & Niger-Congo & Latin & 2400000\\
\hline
 Iban & iba & Autronesian & Latin & 2300000\\
\hline
 Swati & ssw & Niger-Congo & Latin & 2300000\\
\hline
 Fon & fon & Niger-Congo & Latin & 2200000\\
\hline
 Kimbundu & kmb & Niger-Congo & Latin & 2100000\\
\hline
 Acoli & ach & Nilo-Saharan & Latin & 2100000\\
\hline
 Cameroon Pidgin & wes & English Creole & Latin & 2000000\\
\hline
 Urhobo & urh & Niger-Congo & Latin & 2000000\\
\hline
 Lomwe & ngl & Niger-Congo & Latin & 1850000\\
\hline
 Pangasinan & pag & Austronesian & Latin & 1800000\\
\hline
 Latvian & lav & IE: Balto-Slavic & Latin & 1750000\\
\hline
 Alur & alz & Nilo-Saharan & Latin & 1700000\\
\hline
 Aymara & aym & Aymaran & Latin & 1700000\\
\hline
 Batak Toba & bbc & Austronesian & Latin & 1610000\\
\hline
 Wolaytta & wal & Afro-Asiatic & Latin & 1600000\\
\hline
 Sena & seh & Niger-Congo & Latin & 1600000\\
\hline
 Bini & bin & Niger-Congo & Latin & 1600000\\
\hline
 Luba-Katanga & lub & Niger-Congo & Latin & 1505000\\
\hline
 Mende (Sierra Leone) & men & Mande & Latin & 1500000\\
\hline
 Yiddish & yid & IE: Germanic & Aramaic & 1500000\\
\hline
 Cusco Quechua & quz & Quechuan & Latin & 1500000\\
\hline
 Tonga (Zambia) & toi & Niger-Congo & Latin & 1500000\\
\hline
 Kuanyama & kua & Niger-Congo & Latin & 1441000\\
\hline
 Bashkir & bak & Turkic & Cyrillic & 1400000\\
\hline
 Limburgan & lim & IE: Germanic & Latin & 1300000\\
 
\hline
 Southwestern Dinka & dik & Nilo-Saharan & Latin & 1300000\\
\hline
 Venda & ven & Niger-Congo & Latin & 1300000\\
\hline
 Manipuri & mni & Sino-Tibetan & Brahmic & 1250000\\
\hline
 Tswa & tsc & Niger-Congo & Latin & 1200000\\
\hline
 Batak Simalungun & bts & Austronesian & Latin & 1200000\\
\hline
 Sardinian & srd & IE: Italic & Latin & 1175000\\
  
 \hline
\end{tabular}

\end{adjustbox}
\caption{Statistics on all 350 languages present in the MTData sentence retrieval dataset. (3 of 6)} 
\label{tab:all_mtdata_languages_3}
\end{table}

\begin{table}[ht]
\centering
\begin{adjustbox}{width=0.95\textwidth}
\small
\begin{tabular}{||c c c c c||} 
 \hline
 \textbf{Language} & \textbf{ISO-639-3} & \textbf{Language Family} & \textbf{Language Script} & \textbf{Num Native Speakers}\\ [0.5ex] 
 \hline\hline
 
\hline
 Gun & guw & Niger-Congo & Latin & 1162000\\
\hline
 Kekchí & kek & Mayan & Latin & 1100000\\
\hline
 Estonian & est & Uralic & Latin & 1100000\\
\hline
 Zande (individual language) & zne & Niger-Congo & Latin & 1100000\\
\hline
 K'iche' & quc & Mayan & Latin & 1100000\\
\hline
 Morisyen & mfe & French Creole & Latin & 1090000\\
\hline
 Chuvash & chv & Turkic & Cyrillic & 1042989\\
\hline
 Kabiyè & kbp & Niger-Congo & Latin & 1000000\\
\hline
 Songe & sop & Niger-Congo & Latin & 1000000\\
\hline
 Central Huasteca Nahuatl & nch & Uto-Aztecan & Latin & 1000000\\
\hline
 Chokwe & cjk & Niger-Congo & Latin & 980000\\
\hline
 Chuwabu & chw & Niger-Congo & Latin & 970000\\
\hline
 Kachin & kac & Sino-Tibetan & Latin & 940000\\
\hline
 Ayacucho Quechua & quy & Quechuan & Latin & 918200\\
\hline
 Welsh & cym & IE: Celtic & Latin & 892200\\
\hline
 Ngaju & nij & Austronesian & Latin & 890000\\
\hline
 Kabuverdianu & kea & English Creole & Latin & 871000\\
\hline
 Bulu (Cameroon) & bum & Niger-Congo & Latin & 860000\\
\hline
 Lushai & lus & Sino-Tibetan & Brahmic & 843750\\
\hline
 Ndonga & ndo & Niger-Congo & Latin & 810000\\
\hline
 Adangme & ada & Niger-Congo & Latin & 800000\\
\hline
 Yucateco & yua & Mayan & Latin & 770000\\
\hline
 Nias & nia & Austronesian & Latin & 770000\\
\hline
 Chopi & cce & Niger-Congo & Latin & 760000\\
\hline
 Tetela & tll & Niger-Congo & Latin & 760000\\
\hline
 Basque & eus & Basque & Latin & 750000\\
\hline
 Nyaneka & nyk & Niger-Congo & Latin & 750000\\
\hline
 Lozi & loz & Niger-Congo & Latin & 725000\\
\hline
 Chavacano & cbk & IE: Italic & Latin & 700000\\
\hline
 Luvale & lue & Niger-Congo & Latin & 640000\\
\hline
 Konzo & koo & Niger-Congo & Latin & 610000\\
\hline
 Walloon & wln & IE: Italic & Latin & 600000\\
\hline
 Mam & mam & Mayan & Latin & 600000\\
\hline
 Batak Karo & btx & Austronesian & Latin & 600000\\
\hline
 Luxembourgish & ltz & IE: Germanic & Latin & 600000\\
\hline
 Ossetian & oss & IE: Indo-Iranian & Cyrillic & 597450\\
\hline
 Tzeltal & tzh & Mayan & Latin & 590000\\
\hline
 Balkan Romani & rmn & IE: Indo-Iranian & Latin & 563670\\
\hline
 Udmurt & udm & Uralic & Cyrillic & 554000\\
\hline
 Tzotzil & tzo & Mayan & Latin & 550000\\
\hline
 Norwegian Nynorsk & nno & IE: Germanic & Latin & 532000\\
\hline
 Southern Kisi & kss & Niger-Congo & Latin & 530000\\
\hline
 Maltese & mlt & Afro-Asiatic & Latin & 520000\\
\hline
 Samoan & smo & Autronesian & Latin & 510000\\
\hline
 Mambwe-Lungu & mgr & Niger-Congo & Latin & 500000\\
\hline
 Tamashek & tmh & Afro-Asiatic & Latin & 500000\\
\hline
 Krio & kri & English Creole & Latin & 500000\\
\hline
 Imbabura Highland Quichua & qvi & Quechuan & Latin & 500000\\
\hline
 Tooro & ttj & Niger-Congo & Latin & 490000\\
  
 \hline
\end{tabular}

\end{adjustbox}
\caption{Statistics on all 350 languages present in the MTData sentence retrieval dataset. (4 of 6)} 
\label{tab:all_mtdata_languages_4}
\end{table}

\begin{table}[ht]
\centering
\begin{adjustbox}{width=0.95\textwidth}
\small
\begin{tabular}{||c c c c c||} 
 \hline
 \textbf{Language} & \textbf{ISO-639-3} & \textbf{Language Family} & \textbf{Language Script} & \textbf{Num Native Speakers}\\ [0.5ex] 
 \hline\hline
 
\hline
 Western Frisian & fry & IE: Germanic & Latin & 470000\\
\hline
 Sango & sag & French Creole & Latin & 450000\\
\hline
 Plautdietsch & pdt & IE: Germanic & Latin & 450000\\
\hline
 Occitan (post 1500) & oci & IE: Italic & Latin & 450000\\
\hline
 Chimborazo Highland Quichua & qug & Quechuan & Latin & 450000\\
\hline
 Hakha Chin & cnh & Sino-Tibetan & Latin & 446264\\
\hline
 Nyungwe & nyu & Niger-Congo & Latin & 440000\\
\hline
 Friulian & fur & IE: Italic & Latin & 420000\\
\hline
 Isoko & iso & Niger-Congo & Latin & 420000\\
\hline
 Nzima & nzi & Niger-Congo & Latin & 412000\\
\hline
 Catalan & cat & IE: Italic & Latin & 410000\\
\hline
 Kaqchikel & cak & Mayan & Latin & 410000\\
\hline
 Efik & efi & Niger-Congo & Latin & 400000\\
\hline
 Ibanag & ibg & Autronesian & Latin & 400000\\
\hline
 Lunda & lun & Niger-Congo & Latin & 400000\\
\hline
 Tetun Dili & tdt & Autronesian & Latin & 390000\\
\hline
 Gitonga & toh & Niger-Congo & Latin & 380000\\
\hline
 Mingrelian & xmf & Kartvelian & Georgian & 344000\\
\hline
 Papiamento & pap & IE: Italic & Latin & 341300\\
\hline
 Dhivehi & div & IE: Indo-Iranian & Thaana & 340000\\
\hline
 Fijian & fij & Austronesian & Latin & 339210\\
\hline
 Icelandic & isl & IE: Germanic & Latin & 314000\\
\hline
 Wayuu & guc & Arawakan & Latin & 305000\\
\hline
 Esan & ish & Niger-Congo & Latin & 300000\\
\hline
 Basa (Cameroon) & bas & Niger-Congo & Latin & 300000\\
\hline
 Tuvinian & tyv & Turkic & Cyrillic & 280000\\
\hline
 Mapudungun & arn & Araucanian & Latin & 260000\\
\hline
 Ruund & rnd & Niger-Congo & Latin & 250000\\
\hline
 Syriac & syr & Afro-Asiatic & Aramaic & 240000\\
\hline
 Kaonde & kqn & Niger-Congo & Latin & 240000\\
\hline
 Huautla Mazatec & mau & Oto-Manguean & Latin & 240000\\
\hline
 Nyemba & nba & Niger-Congo & Latin & 232000\\
\hline
 Herero & her & Niger-Congo & Latin & 211700\\
\hline
 Breton & bre & IE: Celtic & Latin & 210000\\
\hline
 Amis & ami & Austronesian & Latin & 200000\\
\hline
 Garifuna & cab & Arawakan & Latin & 200000\\
\hline
 Sangir & sxn & Austronesian & Latin & 200000\\
\hline
 Northern Puebla Nahuatl & ncj & Uto-Aztecan & Latin & 200000\\
\hline
 Lamba & lam & Niger-Congo & Latin & 200000\\
\hline
 Abkhazian & abk & Northwest Caucasian & Cyrillic & 190000\\
\hline
 Tonga (Tonga Islands) & ton & Austronesian & Latin & 187000\\
\hline
 Tahitian & tah & Austronesian & Latin & 185000\\
\hline
 Navajo & nav & Dené-Yeniseian & Latin & 170000\\
\hline
 Ngäbere & gym & Chibchan & Latin & 170000\\
\hline
 Irish & gle & IE: Celtic & Latin & 170000\\
\hline
 Tonga (Nyasa) & tog & Niger-Congo & Latin & 170000\\
\hline
 Kwangali & kwn & Niger-Congo & Latin & 152000\\
\hline
 Malinaltepec Me'phaa & tcf & Oto-Manguean & Latin & 150000\\
\hline
 Belize Kriol English & bzj & English Creole & Latin & 150000\\
\hline
 Metlatónoc Mixtec & mxv & Oto-Manguean & Latin & 150000\\
\hline
 Guerrero Nahuatl & ngu & Uto-Aztecan & Latin & 150000\\
\hline
 Purepecha & tsz & Purepecha & Latin & 140000\\
\hline
 Kadazan Dusun & dtp & Autronesian & Latin & 140000\\
\hline
 Sranan Tongo & srn & English Creole & Latin & 130000\\
\hline
 Tok Pisin & tpi & English Creole & Latin & 120000\\
\hline
 Gilbertese & gil & Austronesian & Latin & 120000\\
  
 \hline
\end{tabular}

\end{adjustbox}
\caption{Statistics on all 350 languages present in the MTData sentence retrieval dataset. (5 of 6)} 
\label{tab:all_mtdata_languages_5}
\end{table}

\begin{table}[ht]
\centering
\begin{adjustbox}{width=0.95\textwidth}
\small
\begin{tabular}{||c c c c c||} 
 \hline
 \textbf{Language} & \textbf{ISO-639-3} & \textbf{Language Family} & \textbf{Language Script} & \textbf{Num Native Speakers}\\ [0.5ex] 
 \hline\hline
 
\hline
 Paite Chin & pck & Sino-Tibetan & Latin & 100000\\
\hline
 Saramaccan & srm & English Creole & Latin & 90000\\
\hline
 Duala & dua & Niger-Congo & Latin & 87700\\
\hline
 Isthmus Zapotec & zai & Oto-Manguean & Latin & 85000\\
\hline
 Galela & gbi & West Papuan & Latin & 80000\\
\hline
 Papantla Totonac & top & Mayan & Latin & 80000\\
\hline
 Seselwa Creole French & crs & French Creole & Latin & 73000\\
\hline
 Faroese & fao & IE: Germanic & Latin & 72000\\
\hline
 Lukpa & dop & Niger-Congo & Latin & 70000\\
\hline
 Biak & bhw & Austronesian & Latin & 70000\\
\hline
 Tojolabal & toj & Mayan & Latin & 67000\\
\hline
 Eastern Maroon Creole & djk & English Creole & Latin & 67000\\
\hline
 Guerrero Amuzgo & amu & Oto-Manguean & Latin & 60000\\
\hline
 Chamorro & cha & Autronesian & Latin & 58000\\
\hline
 Scottish Gaelic & gla & IE: Celtic & Latin & 57000\\
\hline
 Kalaallisut & kal & Eskimo–Aleut & Latin & 56000\\
\hline
 Southern Altai & alt & Turkic & Cyrillic & 55720\\
\hline
 Marshallese & mah & Austronesian & Latin & 55000\\
\hline
 Aguaruna & agr & Chicham & Latin & 53400\\
\hline
 Chuukese & chk & Austronesian & Latin & 51330\\
\hline
 Aragonese & arg & IE: Italic & Latin & 50000\\
\hline
 Maori & mri & Austronesian & Latin & 50000\\
\hline
 Coatlán Mixe & mco & Mixe–Zoque & Latin & 45000\\
\hline
 Chol & ctu & Mayan & Latin & 43870\\
\hline
 Inuktitut & iku & Eskimo–Aleut & Inuktitut syllabics & 39770\\
\hline
 Asháninka & cni & Arawakan & Latin & 35000\\
\hline
 Shuar & jiv & Chicham & Latin & 35000\\
\hline
 Pohnpeian & pon & Austronesian & Latin & 29000\\
\hline
 Jakun & jak & Austronesian & Latin & 28000\\
\hline
 Northern Sami & sme & Uralic & Latin & 25000\\
\hline
 Okpe (Southwestern Edo) & oke & Niger-Congo & Latin & 25000\\
\hline
 Pijin & pis & English Creole & Latin & 24000\\
\hline
 Uma & ppk & Austronesian & Latin & 20000\\
\hline
 Northwestern Ojibwa & ojb & Algic & Latin & 20000\\
\hline
 Tena Lowland Quichua & quw & Quechuan & Latin & 17855\\
\hline
 Central Puebla Nahuatl & ncx & Uto-Aztecan & Latin & 16000\\
\hline
 Mirandese & mwl & IE: Italic & Latin & 15000\\
\hline
 Dehu & dhv & Austronesian & Latin & 13000\\
\hline
 Wallisian & wls & Austronesian & Latin & 10400\\
\hline
 Bislama & bis & IE: Germanic & Latin & 10000\\
\hline
 Akawaio & ake & Cariban & Latin & 10000\\
\hline
 Quiotepec Chinantec & chq & Oto-Manguean & Latin & 10000\\
\hline
 Cabécar & cjp & Chibchan & Latin & 8800\\
\hline
 Yapese & yap & Austronesian & Latin & 5130\\
\hline
 Uspanteco & usp & Mayan & Latin & 5100\\
\hline
 Camsá & kbh & Oto-Manguean & Camsa & 4000\\
\hline
 Achuar-Shiwiar & acu & Chicham & Latin & 4000\\
\hline
 Tetelcingo Nahuatl & nhg & Uto-Aztecan & Latin & 3500\\
\hline
 Cherokee & chr & Iroquoian & Latin & 2100\\
\hline
 Asturian & ast & IE: Italic & Latin & 2000\\
\hline
 Niuean & niu & Austronesian & Latin & 2000\\
\hline
 Barasana-Eduria & bsn & Tucanoan & Latin & 1900\\
\hline
 Interlingua (International Auxiliary Language Association) & ina & Constructed & Latin & 1500\\
\hline
 Esperanto & epo & Constructed & Latin & 1000\\
\hline
 Cornish & cor & IE: Celtic & Latin & 557\\
\hline
 Rarotongan & rar & Austronesian & Latin & 450\\
\hline
 Hiri Motu & hmo & Austronesian & Latin & 100\\
\hline
 Potawatomi & pot & Algic & Latin & 100\\
\hline
 Manx & glv & IE: Celtic & Latin & 53\\
\hline
 Klingon & tlh & Constructed & Latin & 25\\
\hline
 Ido & ido & Constructed & Latin & 25\\
\hline
 Volapük & vol & Constructed & Latin & 20\\
\hline
 Latin & lat & IE: Italic & Latin & 1\\
\hline
 Interlingue & ile & Constructed & Latin & 1\\
\hline
 Coptic & cop & Afro-Asiatic & Coptic & 1\\
\hline
 Lojban & jbo & Constructed & Latin & 1\\
\hline
 Lingua Franca Nova & lfn & Constructed & Latin & 1\\
  
 \hline
\end{tabular}

\end{adjustbox}
\caption{Statistics on all 350 languages present in the MTData sentence retrieval dataset. (6 of 6)} 
\label{tab:all_mtdata_languages_6}
\end{table}

\section{Pseudocode for Perturbation}
\label{app:pseudocode_perturbations}

\newcommand{\myalgorithmd}{
\begin{algorithm}[h]
\small
  \SetKw{Kw}{\KwInput}
  \SetKwProg{Fn}{Function}{:}{\KwRet perturbed\_text}
  \SetKwFunction{Func}{NeighborFlip}
  \Fn{\Func{$\rho \leftarrow 0.5$,text$\leftarrow$\texttt{list}}}{

  perturbed\_tokens $\leftarrow$ \texttt{list()}\;
  held\_token $\leftarrow$ \texttt{list(text[0])}
  
  \For{token {\bf in} $\rm{text}[1:]$}{
  %p$\leftarrow$ \texttt{random()}\;
  p $\sim Unif\left(\left[0,1\right]\right)$\;
  \eIf{$p < \rho$}{
  perturbed\_tokens.\texttt{append}(held\_token)\;
  held\_token $\leftarrow$ \texttt{list}(token)
  }{
  perturbed\_tokens $\leftarrow \left[
  \rm{perturbed\_tokens,token} \right]$\;
  }
  }
  perturbed\_tokens.append(held\_token)\;
  perturbed\_text $\leftarrow$ `'.\texttt{join}(perturbed\_tokens)
  }
  \caption{Pseudocode for NeighborFlip.}
\end{algorithm}
}
% }

% \myalgorithma

% \myalgorithmb

% \myalgorithmc

\myalgorithmd

% \begin{figure*}[htb]
%   \null\hfill \myalgorithma \hfill \myalgorithmb \hfill\null\par \medskip
%   \null\hfill \myalgorithma \hfill \myalgorithma \hfill\null
%   \caption{Here are some algorithms.}
%   \end{figure*}

\newpage

\end{document}